\def\year{2021}\relax
\documentclass[letterpaper]{article} 
\usepackage{aaai21}  
\usepackage{times}  
\usepackage{helvet} 
\usepackage{courier}  
\usepackage[hyphens]{url}  
\usepackage{graphicx} 
\usepackage{natbib}  
\usepackage{caption} 
\usepackage[utf8]{inputenc} 
\usepackage{url}            
\usepackage{booktabs}       
\usepackage{amsfonts}       
\usepackage{nicefrac}       
\usepackage{microtype}      
\usepackage[page]{appendix}
\usepackage{amsmath,amssymb}       
\usepackage{bm}
\usepackage{wrapfig}
\usepackage{mathtools}
\usepackage{subcaption}
\usepackage[ruled,vlined,linesnumbered]{algorithm2e}
\usepackage[usenames,dvipsnames]{xcolor}
\usepackage{varwidth}
\usepackage{enumitem}
\usepackage{multirow}
\usepackage[switch]{lineno}

\DeclareCaptionFormat{myformat}{%
  \begin{varwidth}{\linewidth}%
    \centering
    #1#2#3%
  \end{varwidth}%
}

\newcommand \bt {{\boldsymbol{\theta}}}
\newcommand \be {{\boldsymbol{\epsilon}}}

\def\MSHangBox#1{
\begin{minipage}[t]{\textwidth}
\begin{tabbing} 
~\\[-\baselineskip] 
#1 
\end{tabbing}
\end{minipage}} 

\usepackage{footnote}
\newcommand\blfootnote[1]{%
  \begingroup
  \renewcommand\thefootnote{}\footnote{#1}%
  \addtocounter{footnote}{-1}%
  \endgroup
}

\title{Inverse Reinforcement Learning with Explicit Policy Estimates}

\author{
  Navyata Sanghvi$^{\dag}$\textsuperscript{1}, \;
  Shinnosuke Usami$^{\dag}$\textsuperscript{1,2}, \;
  Mohit Sharma$^{\dag}$\textsuperscript{1}, \;
  Joachim Groeger$^*$\textsuperscript{1}, \;
  Kris Kitani\textsuperscript{1} \\ \vspace{2mm}
  {\normalsize{\textsuperscript{1}{Carnegie Mellon University} \hspace{7mm} 
  \textsuperscript{2}{Sony Corporation}}\\}
}

\affiliations{
\small
    \texttt{navyatasanghvi@cmu.edu}, \;\; \texttt{shinnosuke.usami@sony.com}, \;\;
    \texttt{mohitsharma@cmu.edu},\\
    \texttt{jrg@joachimgroeger.com}, \;\; \texttt{kkitani@cs.cmu.edu}
}

\begin{document}
\maketitle

\begin{abstract}
Various methods for solving the inverse reinforcement learning (IRL) problem have been developed independently in machine learning and economics. In particular, the method of Maximum Causal Entropy IRL is based on the perspective of entropy maximization, while related advances in the field of economics instead assume the existence of unobserved action shocks to explain expert behavior (Nested Fixed Point Algorithm, Conditional Choice Probability method, Nested Pseudo-Likelihood Algorithm). In this work, we make previously unknown connections between these related methods from both fields. We achieve this by showing that they all belong to a class of optimization problems, characterized by a common form of the objective, the associated policy and the objective gradient. 
We demonstrate key computational and algorithmic differences which arise between the methods due to an approximation of the optimal soft value function, and describe how this leads to more efficient algorithms. Using insights which emerge from our study of this class of optimization problems, we identify various problem scenarios and investigate each method's suitability for these problems.
\end{abstract}

\blfootnote{\hspace{-7mm} $^{\dag}$Equal contribution. $^*$Work done while at CMU.}

\section{Introduction}

Inverse Reinforcement Learning (IRL) -- the problem of inferring the reward function from observed behavior -- has been studied independently both in machine learning (ML) \cite{abbeel2004apprenticeship, Ratliff2006, boularias2011relative} and economics \cite{miller, pakes, rust_gmc, wolpin}. One of the most popular IRL approaches in the field of machine learning is Maximum Causal Entropy IRL \cite{ziebart_phd}. While this approach is based on the perspective of entropy maximization, independent advances in the field of economics instead assume the existence of unobserved action shocks to explain expert behavior \cite{rust_theory}. Both these approaches optimize likelihood-based objectives, and are computationally expensive. To ease the computational burden, related methods in economics make additional assumptions to infer rewards \cite{hotz, aguirregabiria2002swapping}. While the perspectives these four methods take suggest a relationship between them, to the best of our knowledge, we are the first to make explicit connections between them. 
The development of a common theoretical framework results in a unified perspective of related methods from both fields. This enables us to compare the suitability of methods for various problem scenarios, based on their underlying assumptions and the resultant quality of solutions.


To establish these connections, we first develop a common optimization problem form, and describe the associated objective, policy and gradient forms. We then show how each method solves a particular instance of this common form.
Based on this common form, we show how estimating the optimal soft value function is a key characteristic which differentiates the methods. This difference results in two algorithmic perspectives, which we call optimization- and approximation-based methods.
We investigate insights derived from our study of the common optimization problem towards determining the suitability of the methods for various problem settings. 

Our contributions include: 
\;(1) developing a unified perspective of methods proposed by \citet{ziebart_phd, rust_gmc, hotz, aguirregabiria2002swapping} as particular instances of {\bf a class of IRL optimization problems} that share a common objective and policy form (Section \ref{sec:theory});
\;(2) explicitly demonstrating {\bf algorithmic and computational differences} between methods, which arise from a difference in soft value function estimation (Section \ref{sec:ccp-irl});
\;(3) investigating the {\bf suitability of methods} for various types of problems, using insights which emerge from a study of our unified perspective 
(Section \ref{sec:experiments}).

\section{Related Work}
Many formulations of the IRL problem have been proposed previously, including
maximum margin formulations \cite{abbeel2004apprenticeship, Ratliff2006} and probabilistic formulations \cite{ziebart_phd}. These methods are computationally expensive as they require repeatedly solving the underlying MDP. We look at some methods which reduce this computational burden.

One set of approaches avoids the repeated computation by casting the estimation problem as a supervised classification or regression problem \cite{klein2012inverse, klein2013cascaded}.
Structured Classification IRL (SC-IRL) \cite{klein2012inverse} assumes a linearly parameterized reward and uses expert policy estimates to reduce the IRL problem to a multi-class classification problem. However, SC-IRL is restricted by its assumption of a \emph{linearly} parameterized reward
function.

Another work that avoids solving the MDP repeatedly is Relative Entropy IRL (RelEnt-IRL) \cite{boularias2011relative}. RelEnt-IRL uses a baseline policy for value function approximation. However, such baseline policies are in general not known \cite{ziebart_phd}, and thus RelEnt-IRL cannot be applied in such scenarios.


One method that avoids solving the MDP problem focuses on linearly solvable MDPs \cite{todorov2007linearly}. \cite{dvijotham2010inverse} present an efficient IRL algorithm, which they call OptV, for such linearly solvable MDPs. However, this class of MDPs \emph{assumes} that the Bellman equation can be transformed into a linear equation. 
Also, OptV uses a value-function parameterization instead of a reward-function parameterization, it can have difficulties with generalization when it is not possible to transfer value-function parameters to new environments \cite{ziebart_phd, levine2012continuous}.

Recent work that avoids solving the MDP repeatedly is the CCP-IRL approach \cite{sharma2017inverse}, which observes a connection between Maximum Causal Entropy IRL (MCE-IRL) and Dynamic Discrete Choice models, and uses it to introduce a conditional choice probability (CCP)-based IRL algorithm. On the other hand, our work establishes formal connections between MCE-IRL and a suite of approximation-based methods, of which the CCP method is but one instance. Unlike recent work, we perform a comprehensive theoretical and empirical analysis of each algorithm in the context of trade-offs between the correctness of the inferred solution and its computational burden. 

\section{Preliminaries}
\label{sec:prelims}

In this section,
we first introduce the forward decision problem formulation used in economics literature.
We then familiarize the reader with the inverse problem of interest, \emph{i.e.,} inferring the reward function, and the associated notation.

The Dynamic Discrete Choice (DDC) model is a discrete Markov Decision Process (MDP) with \textit{action shocks}. A DDC is represented by the tuple $(\mathcal{S,A}, T,r,\gamma,\mathcal{E},F)$. $\mathcal{S}$ and $\mathcal{A}$ are a countable sets of states and actions respectively. $T: \mathcal{S} \times \mathcal{A} \times \mathcal{S} \rightarrow [0,1]$ is the transition function. $r: \mathcal{A} \times \mathcal{S} \rightarrow \mathbb{R}$ is the reward function. $\gamma$ is a discount factor. Distinct from the MDP, each action has an associated ``shock'' variable $\epsilon \in \mathcal{E}$, which is unobserved and drawn from a distribution $F$ over $\mathcal{E}$. The vector of shock variables, one for each action, is denoted $\be$. The unobserved shocks $\epsilon_a$ account for agents that sometimes take seemingly sub-optimal actions \cite{mcfadden1973conditional}.
For the rest of this paper, we will use the shorthand $p'$ for transition dynamics $T(s'|s,a)$ and $\text{softmax} f(a) = \exp f(a) / \sum_a \exp f(a)$.

\noindent{\bf The Forward Decision Problem: \;\; }
Similar to reinforcement learning, the DDC forward decision problem in state $(s,\be)$ is to select the action $a$ that maximizes future aggregated utility: $\mathbb{E} \left[\sum_t \gamma^t \left(r(s_t,a_t,\bt) + \epsilon_{a_t}\right) \; | \;(s,\be)\right]$ where the state-action reward function is parametrized by $\bt$. \cite{rust_theory} describes the following Bellman optimality equation for the optimal value function $V^*_\bt(s, \be)$:
\begin{align} 
& V^*_\bt (s, \be) 
\nonumber\\
\label{eq:exantebellman}
& = \textstyle\max\limits_{a\in\mathcal{A}} \big\{ r(s, a, \bt) + \epsilon_{a} + \gamma \mathbb{E}_{s' \sim p', \be'} \left[ V^*_\bt(s', \be') \right]  \big\} .
\end{align}
The optimal choice-specific value is defined as $Q^*_\bt(s,a) \triangleq r(s, a, \bt) + \mathbb{E}_{s' \sim p', \be'} \left[ V^*_\bt(s', \be') \right] $. Then:
\begin{align}
    \label{eqn:v-max-q}
    V^*_\bt(s, \be) = \textstyle\max_{a\in\mathcal{A}} \big\{ \; Q^*_\bt(s,a) + \epsilon_{a} \; \big\}.
\end{align}

\noindent {\bf Solution:\;\;} The choice-specific value $Q^*_\bt(s,a)$ is the fixed point of the contraction mapping $\Lambda_\bt$ \cite{rust_theory}:
\begin{align}
    &\Lambda_\bt (Q)(s,a) 
    \nonumber\\
    \label{eqn:general-contraction}
    & = r(s,a,\bt) + \gamma \mathbb{E}_{s' \sim p', \be'}\left[ \textstyle\max\limits_{a'\in\mathcal{A}} \big\{ Q(s',a') + \epsilon_{a'} \big\} \right].
\end{align}
We denote the indicator function as $\mathbf{I}\{\}$. From (\ref{eqn:v-max-q}), the optimal policy at $(s, \be)$ is given by:
\begin{align}
    \pi(a|s,\be) = \mathbf{I} \left\{a = \textstyle\arg\max_{a'\in\mathcal{A}} \left\{  Q^*_\bt(s,a') + \epsilon_{a'}  \right\} \right\}.
\end{align}

\noindent Therefore, $\pi_\bt(a|s) = \mathbb{E}_\be\left[\pi(a|s,\be)\right]$ is the optimal choice conditional on state alone. $\pi_\bt$ is called the {\emph{conditional choice probability}} (CCP). Notice, $\pi_\bt$ has the same form as a {\emph{policy}} 
in an MDP. 

\noindent{\bf The Inverse Decision Problem: \;\; } The inverse problem, \emph{i.e.}, IRL in machine learning (ML) and structural parameter estimation in econometrics, is to estimate the parameters $\bt$ of a state-action reward function $r(s,a,\bt)$ from expert demonstrations. The expert follows an unknown policy $\pi_E(a|s)$.
A state-action trajectory is denoted: $(\mathbf{s}, \mathbf{a}) = (s_0,a_0, ... , s_T)$. The expert's distribution over trajectories is given by $P_E(\mathbf{s}, \mathbf{a})$. Considering a Markovian environment, the product oNf transition dynamics terms is denoted $P (\mathbf{s}^T || \mathbf{a}^{T-1}) = \prod^T_{\tau=0} P(s_\tau | s_{\tau-1}, a_{\tau-1})$, and the product of  expert policy terms is denoted: $\pi_E(\mathbf{a}^T || \mathbf{s}^T) = \prod^T_{\tau=0} \pi_E(a_\tau |s_\tau) $. The expert distribution is $P_E(\mathbf{s}, \mathbf{a}) = \pi_E(\mathbf{a}^T || \mathbf{s}^T) \; P (\mathbf{s}^T || \mathbf{a}^{T-1})$. Similarly, for a policy $\pi_\bt$ dependent on reward parameters $\bt$, the distribution over trajectories generated using $\pi_\bt$ is given by $P_\bt(\mathbf{s}, \mathbf{a}) = \pi_\bt(\mathbf{a}^T || \mathbf{s}^T) \; P (\mathbf{s}^T || \mathbf{a}^{T-1})$.

\section{A Unified Perspective}
\label{sec:theory}

In order to compare various methods of reward parameter estimation that have been developed in isolation in the fields of economics and ML, it is important to first study their connections and commonality. To facilitate this, in this section, we develop a unified perspective of the following methods:
Maximum Causal Entropy IRL (MCE-IRL) \cite{ziebart_phd}, Nested Fixed Point Algorithm (NFXP) \cite{rust_theory}, Conditional Choice Probability (CCP) \cite{hotz}, and Nested Pseudo-Likelihood Algorithm (NPL) \cite{aguirregabiria2002swapping}. 


To achieve this, we first describe a class of optimization problems that share a common form. While the methods we discuss were derived from different perspectives, we show how each method is a specific instance of this class.  
We characterize this class of optimization problems using a common form of the objective, the associated policy $\pi_\bt$ and the objective gradient. 
In subsequent subsections, we discuss the critical point of difference between the algorithms: \emph{the explicit specification of a policy ${ \tilde{\pi}}$}.

\noindent{\bf Objective Form: \;\; } The common objective in terms of $\bt$ is to maximize expected log likelihood $L(\bt)$ of trajectories generated using a policy $\pi_\bt$, under expert distribution $P_E(\mathbf{s}, \mathbf{a})$:
\begin{align}
    \label{eqn:exp-log-likelihood-full}
    L(\bt) = \mathbb{E}_{P_E(\mathbf{s}, \mathbf{a})}\left[ \log P_\bt(\mathbf{s}, \mathbf{a}) \right].
\end{align}
Since transition dynamics $P (\mathbf{s}^T || \mathbf{a}^{T-1})$ do not depend on $\bt$, maximizing $L(\bt)$ is the same as maximizing $g(\bt)$, \emph{i.e.}, the expected log likelihood of $\pi_\bt(\mathbf{a}^T || \mathbf{s}^T)$ under $P_E(\mathbf{s}, \mathbf{a})$:
\begin{align}
    g(\bt) &= \mathbb{E}_{P_E(\mathbf{s}, \mathbf{a})}\left[ \log \pi_\bt(\mathbf{a}^T || \mathbf{s}^T) \right]
    \nonumber\\
    \label{eqn:exp-log-likelihood-policy}
    &= \mathbb{E}_{P_E(\mathbf{s}, \mathbf{a})}\left[ \textstyle\sum_\tau \log \pi_{\bt}(a_\tau | s_\tau) \right].
\end{align}

\noindent{\bf Policy Form: \;\; } The policy $\pi_\bt$ in objective (\ref{eqn:exp-log-likelihood-policy}) has a general form defined in terms of the state-action ``soft'' value function $Q^{\tilde{\pi}}_\bt(s,a)$ \cite{haarnoja2017reinforcement}, under \emph{some} policy ${ \tilde{\pi}}$.
\begin{align}
    &Q^{\tilde{\pi}}_\bt(s,a) 
    \nonumber\\
    \label{eqn:soft-value}
    &= r(s,a,\bt) + \mathbb{E}_{s', a' \sim \tilde{\pi}} \left[Q^{\tilde{\pi}}_\bt(s',a') - \log \tilde{\pi}(a'|s') \right],\\
    \label{eqn:improved-policy}
    & \pi_\bt(a|s) = \text{softmax } Q^{\tilde{\pi}}_\bt(s,a).
\end{align}
This policy form guarantees that $\pi_\bt$ is an \emph{improvement} over ${ \tilde{\pi}}$ in terms of soft value, \emph{i.e.,} $Q^{\pi_\bt}_\bt(s,a) \geq Q^{\tilde{\pi}}_\bt(s,a), \forall (s,a)$ \cite{haarnoja2018soft}. In subsequent sections, we \emph{explicitly} define ${ \tilde{\pi}}$ in the context of each method. 

\noindent{\bf Gradient Form: \;\; } With this policy form (\ref{eqn:improved-policy}), the gradient of the objective  (\ref{eqn:exp-log-likelihood-policy}) is given by:
\begin{align}
    \textstyle\frac{\partial{g}}{\partial{\bt}} = \textstyle\sum\limits_\tau & \mathbb{E}_{P_E(\mathbf{s}_{1:\tau}, \mathbf{a}_{1:\tau})} \left[
    \textstyle\frac{\partial{Q^{\tilde{\pi}}_\bt(s_\tau, a_\tau)}}{\partial \bt}\right.
    \nonumber\\
    \label{eqn:likelihood-gradient}
    &- \left.\textstyle\sum\limits_{a'} \pi_\bt(a'|s_\tau) \frac{\partial{Q^{\tilde{\pi}}_\bt(s_\tau, a')}}{\partial \bt} \right].
\end{align}
{\emph{Proof:}} Appendix \ref{app:objective-gradient}.

\noindent The general forms we detailed above consider no discounting. In case of discounting by factor $\gamma$, simple modifications apply to the objective, soft value and gradient (\ref{eqn:exp-log-likelihood-policy}, \ref{eqn:soft-value}, \ref{eqn:likelihood-gradient}). 
\\
\emph{Details:} Appendix \ref{app:general-form-discounted}.

\noindent 
We now show how each method is a specific instance of the class of optimization problems characterized by (\ref{eqn:exp-log-likelihood-policy}-\ref{eqn:likelihood-gradient}). Towards this, we explicitly specify $\tilde{\pi}$ in the context of each method. 
We emphasize that, in order to judge how suitable each method is for any problem, it is important to understand the assumptions involved in these specifications and how these assumptions cause differences between methods.



\subsection{Maximum Causal Entropy IRL and Nested Fixed Point Algorithm}
\label{subsec:mce-nfxp}

MCE-IRL \cite{ziebart_phd} and NFXP \cite{rust_theory} originated independently in the ML and economics communities respectively, but they can be shown to be equivalent.
NFXP \cite{rust_theory} solves the DDC forward decision problem for $\pi_\bt$, and maximizes its likelihood under observed data. On the other hand, MCE-IRL formulates the estimation of $\bt$ as the dual of maximizing causal entropy subject to feature matching constraints under the observed data.

\noindent{\bf NFXP: \;\; } Under the assumption that shock values $\epsilon_a$ are \emph{i.i.d} and drawn from a TIEV distribution: $F(\epsilon_a)=e^{-e^{-\epsilon_a}}$, NFXP solves the forward decision problem (Section \ref{sec:prelims}). At the solution, the CCP:
\begin{align}
    \label{eqn:nfxp-policy}
    \pi_\bt(a|s) = \text{softmax } Q^*_{\bt}(s,a),
\end{align} 
where  $Q^*_{\bt}(s,a)$ is the optimal choice-specific value function (\ref{eqn:general-contraction}).
We can show $Q^*_\bt$ is the optimal soft value, and, consequently, $\pi_\bt$ is optimal in the soft value sense.\\\emph{Proof:} Appendix \ref{app:nfxp-proof-pitilde}.

\noindent To estimate $\bt$, NFXP maximizes the expected log likelihood of trajectories generated using $\pi_\bt$ (\ref{eqn:nfxp-policy}) under the expert distribution. We can show the gradient of this objective is:
\begin{align}
    \label{eqn:nfxp-gradient}
    \hspace{-2mm} \textstyle \mathbb{E}_{P_E(\mathbf{s}, \mathbf{a})}\left[\sum\limits_t \frac{\partial r(s_t, a_t, \bt)}{\partial \bt} \right] - \mathbb{E}_{P_\bt(\mathbf{s}, \mathbf{a})}\left[\sum\limits_t \frac{\partial r(s_t, a_t, \bt)}{\partial \bt} \right].
\end{align}
{\emph{Proof:}} Appendix \ref{app:nfxp-gradient}.


\noindent{\bf MCE-IRL: \;} \cite{ziebart_phd} estimates $\bt$ by following the 
dual gradient: 
\begin{align}
    \label{eqn:maxent-gradient}
    \mathbb{E}_{P_E(\mathbf{s}, \mathbf{a})}\left[\textstyle\sum_t \mathbf{f}(s_t, a_t) \right] - \mathbb{E}_{P_\bt(\mathbf{s}, \mathbf{a})}\left[\textstyle\sum_t \mathbf{f}(s_t, a_t) \right],
\end{align} 
where $\mathbf{f}(s,a)$ is a vector of state-action features, and $\mathbb{E}_{P_E(\mathbf{s}, \mathbf{a})}\left[\textstyle\sum_t \mathbf{f}(s_t, a_t) \right]$ is estimated from expert data. The reward is a linear function of features $r(s,a,\bt) = \bt^T \mathbf{f}(s,a)$, and the policy $\pi_\bt(a|s) = \text{softmax} \; Q^{\pi_{\bt}}_{\bt}(s,a)$. This implies $\pi_\bt$ is optimal in the soft value sense \cite{haarnoja2018soft}.

\noindent{\bf Connections: \;\;} From our discussion above we see that, for both NFXP and MCE-IRL, policy $\pi_\bt$ is optimal in the soft value sense. Moreover, when the reward is a linear function of features $r(s,a,\bt) = \bt^T \mathbf{f}(s,a)$, the gradients (\ref{eqn:nfxp-gradient}) and (\ref{eqn:maxent-gradient}) are equivalent. Thus, NFXP and MCE-IRL are equivalent.

\noindent We now show that both methods are instances of the class of optimization problems characterized by (\ref{eqn:exp-log-likelihood-policy}-\ref{eqn:likelihood-gradient}). From the discussion above, $\pi_\bt(a|s) = \text{softmax} \; Q^{\pi_{\bt}}_{\bt}(s,a)$. Comparing this with the forms (\ref{eqn:soft-value}, \ref{eqn:improved-policy}), for these methods, ${ \tilde{\pi}} = \pi_\bt$. Furthermore, by specifying ${ \tilde{\pi}} = \pi_\bt$, we can show that the gradients (\ref{eqn:nfxp-gradient}, \ref{eqn:maxent-gradient}) are equivalent to our objective gradient (\ref{eqn:likelihood-gradient}) (\emph{Proof:} Appendix \ref{app:nfxp-equi-gradient}).
From this we can conclude that both NFXP and MCE-IRL are solving the objective (\ref{eqn:exp-log-likelihood-policy}).

\noindent{\bf Computing $Q^*_\bt$: \;\;} For NFXP and MCE-IRL, every gradient step requires the computation of optimal soft value $Q^*_\bt$. The policy $\pi_\bt$ \eqref{eqn:nfxp-policy} is optimal in the soft value sense. $Q^*_\bt$ is computed using the 
following fixed point iteration. This is a computationally expensive dynamic programming problem.
\begin{align}
    \label{eqn:nfxp-contraction}
    \hspace{-3mm} Q(s,a) \gets & r(s,a,\bt) 
    + \gamma \mathbb{E}_{s' \sim p'}\left[ \log \textstyle\sum_{a'}\exp Q(s', a') \right]
\end{align}


\subsection{Conditional Choice Probability Method 
}
\label{subsec:ccp}
As discussed in Section \ref{subsec:mce-nfxp}, NFXP and MCE-IRL require computing the optimal soft value $Q^*_\bt$ (\ref{eqn:general-contraction}) at every gradient step, which is computationally expensive. To avoid this, the CCP method \cite{hotz} is based on the idea of approximating the optimal soft value $Q^*_\bt$. 
To achieve this, they approximate $Q^*_\bt \approx Q^{\hat{\pi}_E}_\bt$, where $Q^{\hat{\pi}_E}_\bt$ is the soft value under a simple, nonparametric estimate $\hat{\pi}_E$ of the expert's policy $\pi_E$. 
The CCP $\pi_\bt$ \eqref{eqn:nfxp-policy} is then estimated as:
\begin{align}
    \label{eqn:ccp-policy}
    \pi_\bt(a|s) = \text{softmax } Q^{\hat{\pi}_E}_{\bt}(s,a)
\end{align}
In order to estimate parameters $\bt$, the CCP method uses the method of moments estimator \cite{hotz, aguirregabiria2010dynamic}:
\begin{align}
    &\mathbb{E}_{P_E(\mathbf{s}, \mathbf{a})}\left[ \textstyle\sum_\tau \sum_a \mathbf{F}(s_\tau, a)
    \left(\mathbf{I}\{a_\tau = a\} - \pi_\bt(a|s_\tau) \right)
    \right] 
    \nonumber \\
    \label{eqn:ccp-moment}
    &= 0,
\end{align}
where $\mathbf{F}(s,a) = \frac{\partial Q^{\hat{\pi}_E}_\bt(s,a)}{\partial \bt}$ \cite{aguirregabiria2002swapping}. An in-depth discussion of this estimator may be found in \cite{aguirregabiria2010dynamic}. 

\noindent{\bf Connections: \;} We show that CCP is an instance of our class of problems characterized by  (\ref{eqn:exp-log-likelihood-policy}-\ref{eqn:likelihood-gradient}). Comparing \eqref{eqn:ccp-policy} with (\ref{eqn:soft-value}, \ref{eqn:improved-policy}), for this method,  ${ \tilde{\pi}} = \hat{\pi}_E$. Further, by specifying ${ \tilde{\pi}} = {\hat{\pi}_E}$, we obtain $\mathbf{F}(s,a) = \frac{\partial Q^{\tilde{\pi}}_\bt(s,a)}{\partial \bt}$. From \eqref{eqn:ccp-moment}:
\begin{align}
    &\textstyle\sum_\tau \mathbb{E}_{P_E(\mathbf{s}_{1:\tau}, \mathbf{a}_{1:\tau})}\left[\frac{\partial{Q^{\tilde{\pi}}_\bt(s_\tau, a_\tau)}}{\partial \bt} - \sum_{a} \pi_\bt(a|s_\tau) \frac{\partial{Q^{\tilde{\pi}}_\bt(s_\tau, a)}}{\partial \bt} \right] 
    \nonumber\\
    \label{eqn:ccp-gradient}
    &= 0
\end{align}
Notice that this is equivalent to setting our objective gradient (\ref{eqn:likelihood-gradient}) to zero. This occurs at the optimum of our objective (\ref{eqn:exp-log-likelihood-policy}). 

We highlight here that the CCP method is more computationally efficient compared to NFXP and MCE-IRL. At every gradient update step, NFXP and MCE-IRL require \emph{optimizing} the soft value $Q^*_\bt$ \eqref{eqn:nfxp-contraction} to obtain $\pi_\bt$ \eqref{eqn:nfxp-policy}. On the other hand, the CCP method only \emph{improves} the policy $\pi_\bt$ \eqref{eqn:ccp-policy}, which only requires updating the soft value $Q^{\hat{\pi}_E}_\bt$. We show in Section \ref{sec:ccp-irl} how this is more computationally efficient.


\subsection{Nested Pseudo-Likelihood Algorithm 
}
\label{subsec:npl}

Unlike the CCP method which solves the likelihood objective (\ref{eqn:exp-log-likelihood-policy}) once,
NPL is based on the idea of repeated refinement of the approximate optimal soft value $Q^*_\bt$. This results in a refined CCP estimate (\ref{eqn:nfxp-policy}, \ref{eqn:ccp-policy}), and, thus, a refined objective \eqref{eqn:exp-log-likelihood-policy}. NPL solves the objective repeatedly, and its first iteration is equivalent to the CCP method. The authors \cite{aguirregabiria2002swapping} prove that this iterative refinement converges to the NFXP \cite{rust_theory} solution, as long as the first estimate ${ \tilde{\pi}} = \hat{\pi}_E$ is ``sufficiently close'' to the true optimal policy in the soft value sense. We refer the reader to \cite{kasahara2012sequential} for a discussion on convergence criteria. 

\noindent{\bf Connections: \;\;} The initial policy ${ \tilde{\pi}}^1 = \hat{\pi}_E$ is estimated from observed data. Subsequently, ${ \tilde{\pi}}^k = \pi^{k-1}_{\bt^*}$, where $\pi^{k-1}_{\bt^*}$ is the CCP under optimal reward parameters $\bt^*$ from the $k-1^{\text{th}}$ iteration. We have discussed that NPL is equivalent to repeatedly maximizing the objective (\ref{eqn:exp-log-likelihood-policy}), where expert policy $\hat{\pi}_E$ is explicitly estimated, and CCP $\pi^k_\bt$ (\ref{eqn:improved-policy}) is derived from refined soft value approximation $Q^{\tilde{\pi}^k}_\bt \approx Q^*_\bt$. 

\subsection{\bf Summary} 

\noindent{\bf Commonality:} In this section, we  demonstrated connections between reward parameter estimation methods, by developing a common class of optimization problems characterized by general forms (\ref{eqn:exp-log-likelihood-policy}-\ref{eqn:likelihood-gradient}). Table \ref{tab:summary} summarizes this section, with specifications of ${ \tilde{\pi}}$ for each method, and the type of computation required at every gradient step.

\begin{table}[htbp!]
    \centering
    \caption{Summary of the common objective and policy form, and specifications for each method.}
\scalebox{0.72}{
    \begin{tabular}{c|c|c|c}
    \toprule
     \multicolumn{4}{c}{\normalsize Objective Form: $g(\bt) = \mathbb{E}_{P_E(\mathbf{s}, \mathbf{a})}\left[ \textstyle\sum_\tau \log \pi_{\bt}(a_\tau | s_\tau) \right]$}\\
    \multicolumn{4}{c}{\normalsize Policy Form: $\pi_\bt(a|s) = \text{softmax } Q_\bt^{{\tilde{\pi}}}(s,a)$. \; $Q^{{\tilde{\pi}}}_\bt$ is soft value \eqref{eqn:soft-value}}\\
    \toprule
    \bf Method $\rightarrow$ & MCE-IRL & \multirow{2}{*}{CCP Method} & \multirow{2}{*}{NPL}\\
    \bf Characteristic $\downarrow$ & $=$ NFXP & & \\
    \toprule
    \bf Specification of \large$\tilde{\pi}$ &\large $\pi_\bt$ &\large  $\hat{\pi}_E$ & \large $\hat{\pi}_E, \;\;\tilde{\pi} = \; \pi_{\bt^*} \; \pi^2_{\bt^*}, ...$\\
    \midrule
    \bf Gradient step & Soft Value & Policy & Policy \\
    \bf computation & optimization & improvement & improvement\\
    \bottomrule
    \end{tabular}
}
\label{tab:summary}
\end{table}

\noindent{\bf Differences:} In Section \ref{subsec:mce-nfxp}, we discussed that the computation of the NFXP (or MCE-IRL) gradient \eqref{eqn:nfxp-gradient} involves solving the forward decision problem \emph{exactly} in order to correctly infer the reward function. This requires computing a policy ${ \tilde{\pi}} = \pi_\bt$ that is \emph{optimal} in the soft value sense. On the other hand, we discussed in Sections \ref{subsec:ccp}-\ref{subsec:npl} that the computation of the CCP and NPL gradient involves an \emph{approximation} of the optimal soft value. This only requires computing the policy $\pi_\bt$ that is an \emph{improvement} over ${ \tilde{\pi}}$ in the soft value sense. This insight lays the groundwork necessary to compare the methods. The approximation of the soft value results in algorithmic and computational differences between the methods, which we make explicit in Section \ref{sec:ccp-irl}.
Approximating the soft value results in a trade-off between correctness of the inferred solution and its computational burden. The implication of these differences (\emph{i.e.,} approximations) on the suitability of each method is discussed in Section \ref{sec:experiments}.

\section{An Algorithmic Perspective}
\label{sec:ccp-irl}

In this section, we explicitly illustrate the algorithmic differences that arise due to differences in the computation of the soft value function. The development of this perspective is important for us to demonstrate how, as a result of approximation, NPL has a more computationally efficient reward parameter update compared to MCE-IRL.  

\noindent{\bf Optimization-based Methods: \;\; } As described in Section \ref{subsec:mce-nfxp}, NFXP and MCE-IRL require the computation of the optimal soft value $Q^*_\bt$ \eqref{eqn:nfxp-contraction}. Thus, we call these approaches ``optimization-based methods'' and describe them in Alg. \ref{algo:optimization}. We define the future state occupancy for $s'$ when following policy $\pi$: $\text{Occ}^\pi(s') = \sum_t P^\pi\left(s_t = s'\right)$. The gradient \eqref{eqn:nfxp-gradient} can be expressed in terms of occupancy measures:
\begin{align}
    \label{eqn:algo-expdr}
    \boldsymbol{\mu}^{\hat{\pi}_E} &= \textstyle \sum\limits_{s'}\text{\small Occ}^{\hat{\pi}_E}(s') \mathbb{E}_{a' \sim \hat{\pi}_E}\left[\frac{\partial r(s',a',\bt)}{\partial \bt}\right], \\
    \boldsymbol{\mu}^{\pi_\bt} &= \textstyle \sum\limits_{s'}\text{\small Occ}^{\pi_\bt}(s') \mathbb{E}_{a' \sim \pi_\bt}\left[\frac{\partial r(s',a',\bt)}{\partial \bt}\right]\\
    \label{eqn:algo-expdr-grad}
    \textstyle \frac{\partial{g}}{\partial{\bt}} &= \boldsymbol{\mu}^{\hat{\pi}_E} - \boldsymbol{\mu}^{\pi_\bt}
\end{align}

\noindent{\bf Approximation-based Methods: \;\;} As described in Sections (\ref{subsec:ccp}, \ref{subsec:npl}), CCP and NPL avoid optimizing the soft value by approximating $Q^*_\bt \approx Q^{\tilde{\pi}}_\bt$ using a policy $\tilde{\pi}$. We call these approaches ``approximation-based methods'' and describe them in Algorithm \ref{algo:approx}. Note, $K=1$ is the CCP Method.
We define the future state occupancy for $s'$ when beginning in $(s, a)$ and following policy  $\pi$ : $\text{Occ}^\pi(s' | s, a) = \sum_t P^\pi\left(s_t = s' \;|\; (s_0,a_0) = (s,a)\right)$. 
 (\ref{eqn:soft-value}, \ref{eqn:likelihood-gradient}) can be written in terms of occupancy measures as follows:
\begin{align}
    Q^{\tilde{\pi}}_\bt(s,a) =&\; r(s, a, \bt) + \textstyle \sum_{s'} \text{Occ}^{\tilde{\pi}}(s' | s, a) \mathbb{E}_{a' \sim \tilde{\pi}}\big[
    \nonumber\\
    \label{eqn:algo-soft-value}
    &\; \qquad r(s', a', \bt) - \log \tilde{\pi}(a'|s') \big],
    \\
    \textstyle \frac{\partial Q^{\tilde{\pi}}_\bt(s,a)}{\partial \bt} =&\;  \textstyle \frac{\partial r(s,a,\bt)}{\partial \bt} 
    \nonumber\\
    \label{eqn:algo-value-gradient}
    &\;+ \textstyle \sum_{s'}\text{Occ}^{\tilde{\pi}}(s' | s, a) \mathbb{E}_{a' \sim \tilde{\pi}}\left[\frac{\partial r(s',a',\bt)}{\partial \bt}\right]
    \\
    \textstyle \frac{\partial{g}}{\partial{\bt}} =&\; \textstyle \sum_{s'}\text{Occ}^{\hat{\pi}_E}(s') \Big( \mathbb{E}_{a' \sim \hat{\pi}_E}\left[\frac{\partial{Q^{\tilde{\pi}}_\bt(s', a')}}{\partial \bt}\right] 
    \nonumber\\
    \label{eqn:algo-gradient}
    &\; \qquad - \mathbb{E}_{a' \sim \pi_\bt}\left[\textstyle\frac{\partial{Q^{\tilde{\pi}}_\bt(s', a')}}{\partial \bt}\right] \Big).     
\end{align}

 \begin{center}
    \begin{algorithm}
    \DontPrintSemicolon
    \SetAlgoLined
    \KwIn{Expert demonstrations.}
    \KwResult{Reward params. $\bt^*$, policy $\pi_{\bt^*}$.}
     
     Estimate expert policy $\hat{\pi}_E$.\\
     {\bf Evaluate} Occ$^{\hat{\pi}_E}(s') \; \forall s'$. \\
        \Repeat(  { \bf (update reward)}){$\bt$ \text{not converged}}{
            {\bf Optimize} soft value $Q^*_\bt$. \hfill \eqref{eqn:nfxp-contraction}.\\ 
            Compute $\pi_\bt = \text{softmax } Q^*_\bt(s,a)$. \hfill (\ref{eqn:nfxp-policy}).\\
            {\bf Evaluate} $\boldsymbol{\mu}^{\pi_\bt}$. \hfill \eqref{eqn:algo-expdr}.\\
            Update gradient $\frac{\partial g}{\partial \bt}$. \hfill \eqref{eqn:algo-expdr-grad}.\\
            Update $\bt \gets \bt + \alpha \frac{\partial g}{\partial \bt}$ .\\
        }
        $\pi_{\bt^*} \gets \pi_\bt$,
        $\;\;\bt^* \gets \bt$.
     \caption{Optimization-based Method}
     \label{algo:optimization}
    \end{algorithm}
    \vspace{-10mm}
\end{center}
\begin{center}
    \begin{algorithm}
    \DontPrintSemicolon
    \SetAlgoLined
    \KwIn{Expert demonstrations.}
    \KwResult{Reward params. $\bt^*$, policy $\pi_{\bt^*}$.}
     
     Estimate expert policy $\hat{\pi}_E$.\\
     Initialize $\tilde{\pi} \gets \hat{\pi}_E$.\\
     {\bf Evaluate} Occ$^{\hat{\pi}_E}(s') \; \forall s'$. \\
     \For( {\bf (update policy)}){k = 1 ... K}{
        {\bf Evaluate} Occ$^{\tilde{\pi}}(s'|s,a) \; \forall(s',s,a)$.\\
        \Repeat(  { \bf (update reward)}){$\bt$ \text{not converged}}{
            Update value $Q^{\tilde{\pi}}_\bt$. \hfill \eqref{eqn:algo-soft-value}.\\
            Improve $\pi_\bt = \text{softmax } Q^{\tilde{\pi}}_\bt(s,a)$. \hfill (\ref{eqn:improved-policy}).\\
            Update $\frac{\partial Q^{\tilde{\pi}}_\bt}{\partial \bt}$. \hfill \eqref{eqn:algo-value-gradient}.\\
            Update gradient $\frac{\partial g}{\partial \bt}$. \hfill \eqref{eqn:algo-gradient}.\\
            Update $\bt \gets \bt + \alpha \frac{\partial g}{\partial \bt}$ .\\
        }
        {
        $\tilde{\pi} \gets \pi_\bt$,
        $\;\;\;\;\pi_{\bt^*} \gets \pi_\bt$,
        $\;\;\bt^* \gets \bt$.
        }
     }
     \caption{Approximation-based method}
     \label{algo:approx}
    \end{algorithm}
    \vspace{-10mm}
\end{center}

\noindent{\bf Reward Update: \;\; } NPL has a very efficient reward parameter update (\emph{i.e.}, inner) loop (Alg. \ref{algo:approx}: Lines 6-12), compared to the update loop of MCE-IRL (Alg. \ref{algo:optimization}: Lines 3-9). Each gradient step in MCE-IRL (Alg. \ref{algo:optimization}) involves expensive dynamic programming for:\;\; (1) optimizing soft value (Line 4, \eqref{eqn:nfxp-contraction}), and\;\; (2) evaluating $\boldsymbol{\mu}^{\pi_\bt}$ by computing occupancy measures $\text{Occ}^{\pi_\bt}(s')$ (Line 6, \eqref{eqn:nfxp-contraction}). On the other hand, each gradient step in NPL (Alg. \ref{algo:approx}) only involves:\;\; (1) updating soft value $Q^{\tilde{\pi}}_\bt$ (Line 7, \eqref{eqn:algo-soft-value}), and\;\; (2) updating value gradient (Line 9, \eqref{eqn:algo-value-gradient}). Both steps can be efficiently performed without dynamic programming, as the occupancy measures $\text{Occ}^{\tilde{\pi}}(s' | s, a)$ can be pre-computed (Line 5). The value and value gradient (\ref{eqn:algo-soft-value}, \ref{eqn:algo-value-gradient}) are linearly dependent on reward and reward gradient respectively, and can be computed in one step using matrix multiplication. This point is elaborated in Appendix \ref{app:B}. The gradient update step in both algorithms (Alg. \ref{algo:optimization}: Line 7, Alg. \ref{algo:approx}: Line 10) has the same computational complexity, \emph{i.e.,} linear in the size of the environment.

The outer loop in NPL (Alg. \ref{algo:approx}: Lines 4-14) converges in very few iterations ($< 10$) \cite{aguirregabiria2002swapping}. Although computing occupancy measures $\text{Occ}^{\tilde{\pi}}(s' | s, a)$ (Line 5) requires dynamic programming, the number of outer loop iterations is many order of magnitudes fewer than the number of inner loop iterations. Since Alg. \ref{algo:approx} avoids dynamic programming in the inner reward update loop, approximation-based methods are much more \emph{efficient} than optimization-based methods (Alg. \ref{algo:optimization}). The comparison of computational load is made explicit in Appendix \ref{app:B}.


\begin{figure*}[h]
    \centering
    \includegraphics[width=1.0\linewidth]{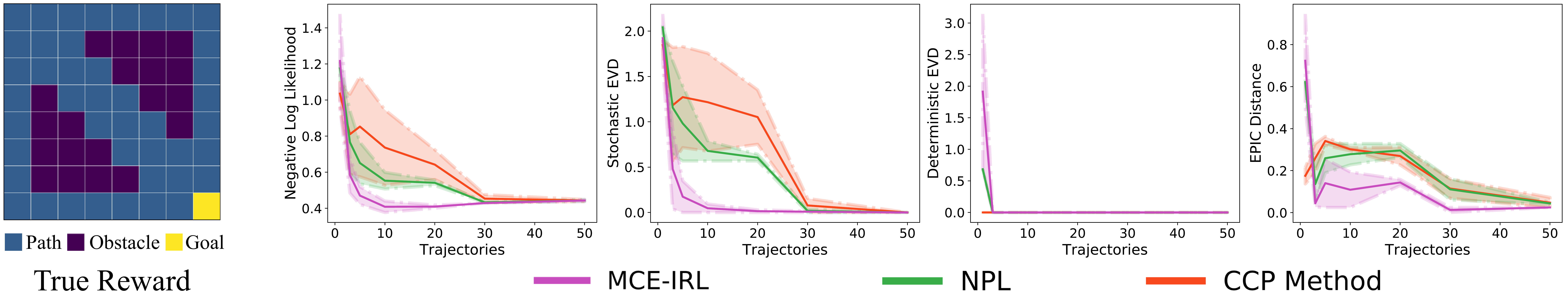}
    \caption{ \footnotesize{\bf Obstacleworld:} Visualization of the true reward and comparative performance results (Performance metrics vs. trajectories).
    }
    \label{fig:img_performance_obstacleworld}
\end{figure*}

\section{Suitability of Methods}\label{sec:experiments}


In Section \ref{sec:ccp-irl}, we made explicit the computational and algorithmic differences between optimization (MCE-IRL, NFXP) and approximation-based (CCP, NPL) methods. While approximation-based methods outperform optimization-based methods in terms of computational efficiency, the approximation the soft value introduces a trade-off between the correctness of the inferred solution and its computational burden. For some types of problems, trading the quality of the inferred reward for computational efficiency 
is unreasonable, so optimization-based methods are more suitable. 
Using theory we developed in Section \ref{sec:theory}, in this section, we develop hypotheses about the hierarchy of methods in various types of problem situations, and investigate each hypothesis using an example.
We quantitatively compare the methods using the following metrics (lower values are better):
\begin{itemize}[leftmargin=*, topsep=0pt, partopsep=0pt]
    \item \textbf{Negative Log Likelihood (NLL)}  evaluates the likelihood of the expert path under the predicted policy $\pi_\bt*$, and is directly related to our objective~\eqref{eqn:exp-log-likelihood-full}.
    \item \textbf{Expected Value Difference (EVD)} is value difference of two policies under {true} reward: 1) optimal policy under {true} reward and 2) optimal policy under \emph{output} reward $\bt^*$. 
    \item \textbf{Stochastic EVD} is the value difference of the following policies under \emph{true} reward: 1) optimal policy under the true reward and 2) the \emph{output} policy $\pi_\bt*$.
    While a low Stochastic EVD may indicate a better output policy $\pi_{\bt^*}$, low EVD may indicate a better output reward $\bt^*$.
    
    \item \textbf{Equivalent-Policy Invariant Comparison (EPIC)} \cite{gleave2020quantifying} is a recently developed metric that measures the distance between two reward functions without training a policy. EPIC is shown to be invariant on an equivalence class of reward functions that always induce the same optimal policy. The EPIC metric $\in [0, 1]$ with lower values indicates similar reward functions. 
\end{itemize}
\noindent EVD and EPIC evaluate inferred reward parameters $\bt^*$, while NLL and Stochastic-EVD evaluate the inferred policy $\pi_{\bt^*}$. In addition to evaluating the recovered reward, evaluating $\pi_{\bt^*}$ is important because different rewards might induce policies that perform similarly well in terms of our objective \eqref{eqn:exp-log-likelihood-policy}. Details for all our experiments are in Appendices \ref{app:C}-\ref{app:D}.

\noindent {\bf Method Dependencies:} From Section \ref{sec:theory}, we observe that the MCE-IRL gradient \eqref{eqn:maxent-gradient} depends on the expert policy estimate $\hat{\pi}_E$ only through the expected \emph{feature count} estimate $\mathbb{E}_{\hat{P}_E(\mathbf{s}, \mathbf{a})}\left[\textstyle\sum_t \mathbf{f}(s_t, a_t) \right]$. In non-linear reward settings, the dependence is through the expected \emph{reward gradient} estimate $\mathbb{E}_{\hat{P}_E(\mathbf{s}, \mathbf{a})}\left[\sum_t  {\partial r(s_t, a_t, \bt)}/{\partial \bt} \right]$. On the other hand, the NPL (or CCP) gradient \eqref{eqn:ccp-gradient} depends on the expert policy estimate for estimating a state's \emph{importance relative} to others (\emph{i.e.,}  state occupancies under the estimated expert policy), and for \; \emph{approximating} the optimal soft value. 

From these insights, we see that the suitability of each method for a given problem scenario depends on: \;(1) the amount of expert data, and on \;(2) how ``good'' the resultant estimates and approximations are in that scenario.
In the following subsections, we introduce different problem scenarios, each characterized by the goodness of these estimates, and investigate our hypotheses regarding each method's suitability for those problems.

\begin{figure*}[tb]
    \centering
    \includegraphics[width=1.0\linewidth]{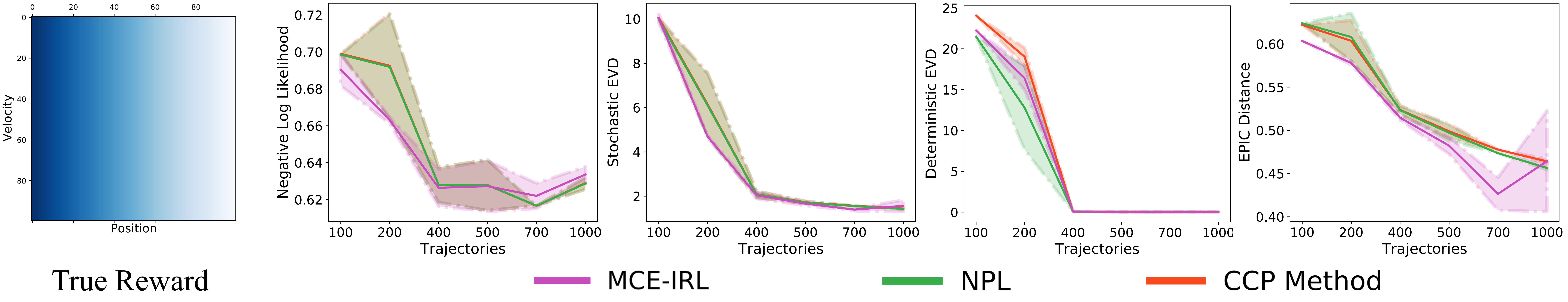}
    \caption{ \footnotesize{\bf Mountain Car:} 
    Visualization of the true reward and comparative performance results (Performance metrics vs. trajectories). 
    }
    \label{fig:img_performance_mountaincar}
\end{figure*}

\begin{figure*}[htbp!]
    \centering
    \includegraphics[width=1.0\textwidth]{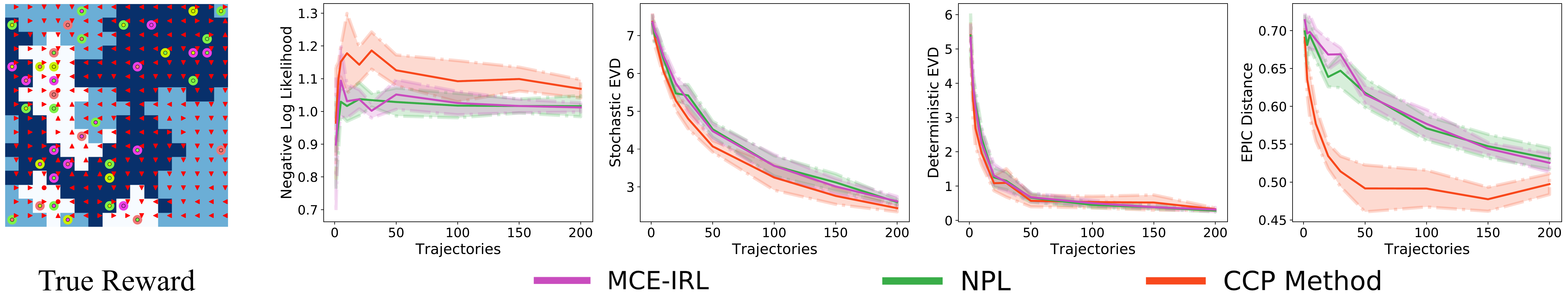}
    \caption{  \footnotesize{{\bf Objectworld:} Visualization of the true reward in $16^{2}$ grid, 4 colors and comparative performance results (Performance metrics vs. trajectories in $32^{2}$ grid, 4 colors). Dark - low reward, Light - high reward. 
    Red arrows represent optimal policy under the reward.
    }}
    \label{fig:img_performance_objectworld}
\end{figure*}

\begin{figure}[ht]
    \centering
    \includegraphics[width=1.0\columnwidth]{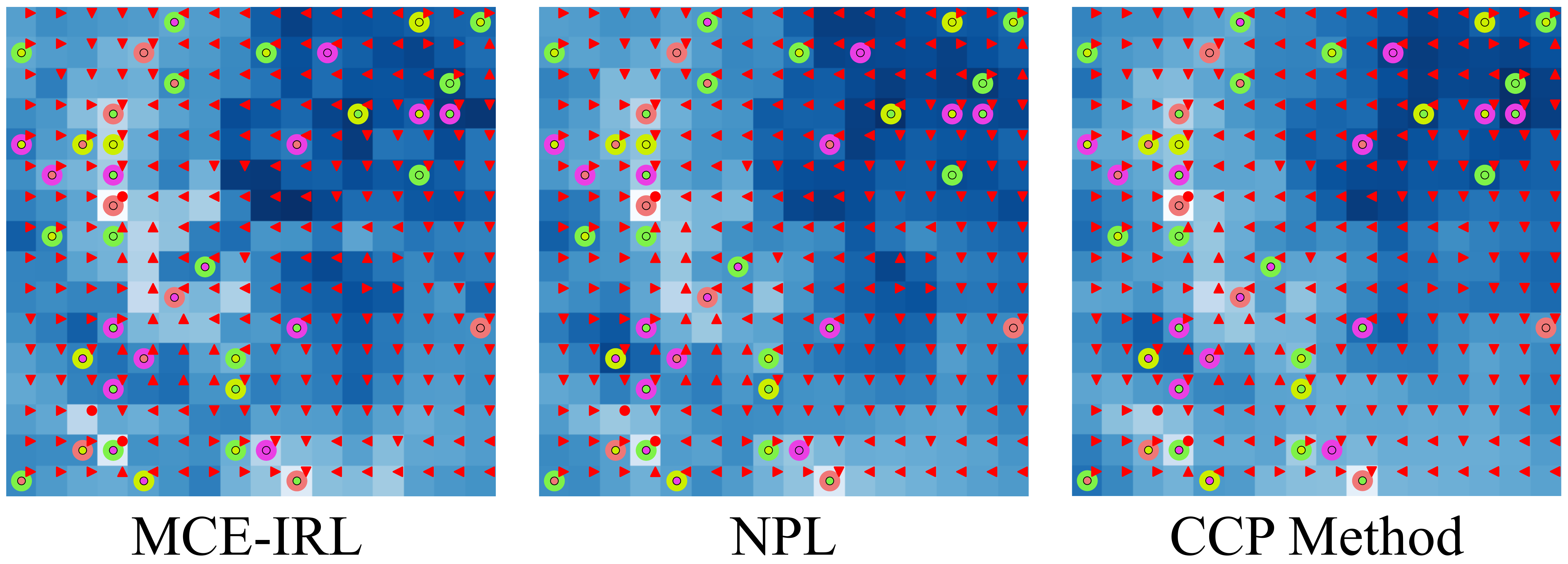}
    \caption{  \footnotesize{{\bf Objectworld:} Recovered reward in $16^{2}$ grid, 4 colors with 200 expert trajectories as input. 
    }}
    \label{fig:img_performance_objectworld_qualitative}
\end{figure}

\subsection{Well-Estimated Feature Counts}\label{subsec:feat}

\noindent {\bf Scenario: \;\;} 
We first investigate scenarios where feature counts can be estimated well even with little expert data. In such scenarios, the feature representation allows distinct states to be correlated. For example, the expert's avoidance of \emph{any} state with obstacles 
should result in identically low preference for \emph{all} states with obstacles. Such a scenario would allow expert feature counts to be estimated well, even when expert data only covers a small portion of the state space.

\noindent {\bf Hypothesis:\;\;} 
Optimization-based methods will perform better than approximation-based methods in low data regimes, and converge to similar performance in high data regimes.

\noindent {\bf Reasons:\;\;} If feature counts can be estimated well from small amounts of data, MCE-IRL (= NFXP) is expected to converge to the correct solution. This follows directly from method dependencies outlined above. On the other hand, NPL (and CCP) require good estimates of a state's relative importance and soft value in order to perform similarly well. Since low data regimes do not allow these to be well-estimated, approximation-based methods are expected to perform as well as optimization-based ones only in high data regimes.

\noindent {\bf Experiment: \;\;}
We test our hypothesis using the Obstacleworld environment  (Figure~\ref{fig:img_performance_obstacleworld}). 
We use three descriptive feature representations \emph{(path, obstacle, goal)} for our states.
Since this representation is simultaneously informative for a \emph{large} set of states, 
we can estimate feature counts well even with little expert data.
The true reward is a linear function of features $(path:\; 0.2, obstacle:\; 0.0, goal:\; 1.0)$.

\noindent 
In Figure~\ref{fig:img_performance_obstacleworld}, we observe that in low data-regimes (i.e. with few trajectories) MCE-IRL performs well on all metrics. However, with low expert data, CCP and NPL perform poorly (\emph{i.e.,} high NLL, Stochastic EVD, EPIC). 
With more expert data, CCP method and NPL converge to similar performance as MCE-IRL. This is in agreement with our hypothesis.


\subsection{Correlation of Feature Counts and Expert Policy Estimates}\label{subsec:feat_not_well_estimated}

\noindent {\bf Scenario: \;\;} 
We now investigate the scenario where the goodness of feature count and expert policy estimates becomes correlated. In other words, a high amount of expert data is required to estimate feature counts well. This scenario may arise when feature representations either (1) incorrectly discriminate between states, or (2) are not informative enough to allow feature counts to be estimated from little data. 

\noindent {\bf Hypothesis:\;\;} 
Both optimization-based and approximation-based methods will perform poorly in low expert data regimes, and do similarly well in high expert data regimes.

\noindent {\bf Reasons:\;\;} 
In this scenario, the goodness of feature count, relative state importance and optimal soft value estimates is similarly dependent on the amount of expert data. Thus we expect, optimization- and approximation-based methods to perform poorly in low data regimes, and similarly well in high data regimes.

\noindent {\bf Experiment: \;\;}
We investigate our hypothesis in the MountainCar environment, with a feature representation that discriminates between all states. Thus, state features are defined as one-hot vectors.
MountainCar is a continuous environment where the state is defined by the position and velocity of the car. The scope of this work is limited to discrete settings with known transition dynamics. Accordingly, we estimate the transition dynamics from continuous expert trajectories using kernels, and discretize the state space to a large set of states ($10^4$).
We define the true reward as distance to the goal.

\noindent In Figure~\ref{fig:img_performance_mountaincar}, we observe that in low-data regimes all methods perform poorly with high values for all metrics. As the amount of expert data increases, the performance of each method improves. More importantly, around the same number of trajectories ($\approx 400$) all methods perform equally well, with a similar range of values across all metrics. This is in agreement with our hypothesis.

\subsection{Deep Reward Representations}
\label{subsec:nonlin}

\noindent {\bf Scenario: \;\;}  
We investigate scenarios where rewards are deep neural network representations of state-action, as opposed to linear representations explored in previous subsections. 

\noindent {\bf Hypothesis:\;\;} 
Optimization-based methods either perform better or worse than the approximation-based methods in low data regimes and perform similarly well in high data regimes.

\noindent {\bf Reasons:\;\;} 
From \eqref{eqn:nfxp-gradient}, the gradient of optimization-based methods depends on the expert policy estimate through the expected reward gradient. Comparing \eqref{eqn:nfxp-gradient} and \eqref{eqn:maxent-gradient}, we can think of the vector of reward gradients ${\partial r(s_t, a_t, \bt)}/{\partial \bt}$ as the state-action feature vector. During learning, since this feature vector is dependent on the current parameters $\bt$, the statistic $\mathbb{E}_{\hat{P}_E(\mathbf{s}, \mathbf{a})}\left[\sum_t  {\partial r(s_t, a_t, \bt)}/{\partial \bt} \right]$ is a non-stationary target in the MCE-IRL gradient. In the low data regime, at every gradient step, this could either be well-estimated (similar to Section \ref{subsec:feat}) or not (similar to Section \ref{subsec:feat_not_well_estimated}), depending on the capacity and depth of the network. On the other hand, in high data regimes, we can expect reward gradient, relative state importance and soft value to all be estimated well, since the expert policy can be estimated well.

\noindent {\bf Experiment: \;\;} 
We investigate the hypothesis using the Objectworld environment \cite{wulfmeier2015maximum} which consists of a non-linear feature representation. 
Objectworld consists of an $N^{2}$ grid and randomly spread through the grid are objects, each with an inner and outer color chosen from $C$ colors.
The feature vector is a set of continuous values $x \in \mathbb{R}^{2C}$, where $x_{2i}$ and $x_{2i+1}$ are state's distances from the \emph{i'th} inner and outer color. 

\noindent From Figure~\ref{fig:img_performance_objectworld}, in low-data regimes, all methods perform poorly with high values for all metrics. With more expert data, the performance for all methods improve and converge to similar values. Similar results for MCE-IRL were observed in \cite{wulfmeier2015maximum}. Consistent with our hypothesis, in this environment we observe no difference in the performance of optimization- and approximation-based methods in both low and high data regimes.


\subsection{Discussion}

In Sections \ref{subsec:feat}-\ref{subsec:nonlin}, we described three problem scenarios and discussed the performance of optimization- and approximation-based methods. We now discuss the suitability of methods for these scenarios.

\noindent {\bf High data regimes: \;\;} In all scenarios we discussed, in high data regimes, both optimization- and approximation-based methods perform similarly well on all metrics (Figures \ref{fig:img_performance_obstacleworld}-\ref{fig:img_performance_objectworld}). Qualitatively, all methods also recover similar rewards in high data regimes (Figures \ref{fig:img_performance_objectworld_qualitative}, 7 (Appendix \ref{app:D})). This is because, as stated in Section~\ref{sec:theory}, NPL converges to NFXP when the expert policy is well-estimated, \emph{i.e.,}, when more data is available. Further, approximation-based methods are significantly more computationally efficient than optimization-based methods (Section \ref{sec:ccp-irl}). This finding is empirically supported in Table \ref{table:table_time_result}. From these observations, we conclude that approximation-based methods are more suitable than optimization-based methods in high data regimes.

\noindent{\bf Low data regimes: \;\;} In Sections \ref{subsec:feat_not_well_estimated}-\ref{subsec:nonlin}, we introduced two scenarios where approximation- and optimization-based methods both perform similarly (poorly) in low-data regimes (Figures \ref{fig:img_performance_mountaincar}, \ref{fig:img_performance_objectworld}). Since approximation-based methods always outperform optimization-based methods in terms of computational efficiency (Table \ref{table:table_time_result}), in these scenarios, approximation-based methods are more suitable. On the other hand, optimization-based methods are more suitable when feature counts can be estimated well from little data (Section \ref{subsec:feat}).

\begin{table}[t]
    \centering
    \resizebox{1.0\columnwidth}{!}{%
    \begin{tabular}{@{}lrrr@{}}
    \toprule
    {\bf Settings} & MCE-IRL & NPL   & CCP Method \\ \midrule

    \multicolumn{4}{c}{MountainCar} \\ \midrule
    State Size: 100$^{2}$   & 4195.76 \;    & 589.21  ($\times$7)  & 193.95 ($\times$\textbf{22}) \\ \midrule 
    \multicolumn{4}{c}{ObjectWorld} \\ \midrule
    Grid: 16$^{2}$, C: 4 & 32.21   & 4.18 \; ($\times$8)   & 2.40  ($\times$ \textbf{13})    \\
    Grid: 32$^{2}$, C: 2 & 638.83  & 30.63  ($\times$21) & 14.00  ($\times$ \textbf{46})   \\
    Grid: 32$^{2}$, C: 4 & 471.64  & 29.85 ($\times$16) & 11.19 ($\times$ \textbf{42})   \\ 
    Grid: 64$^{2}$, C: 4 & 10699.47  & 340.55 ($\times$31) & 103.79 ($\times$ \textbf{103})   \\ \bottomrule
    \end{tabular}
    }
    \caption{\footnotesize{\bf Training time} (seconds) averaged across multiple runs. Numbers in brackets correspond to the speed up against MCE-IRL. }
    \label{table:table_time_result} 
\end{table}

\subsection{Conclusions}
In this work, we explicitly derived connections between four methods of reward parameter estimation developed independently in the fields of economics and ML. To the best of our knowledge, we are the first to bring these methods under a common umbrella. We achieved this by deriving a class of optimization problems, of which each method is a special instance. We showed how a difference in the estimation of the optimal soft value results in different specifications of the explicit policy $\tilde{\pi}$, and used our insights to demonstrate algorithmic and computational differences between methods. 
Using this common form we analyzed the applicability of each method in different problem settings. Our analysis shows how approximation-based methods are superior to optimization-based methods in some settings and vice-versa.

Additionally, approximation-based approaches have been applied to situations with continuous state or action spaces \cite{altuug1998effect}.
Such settings are outside of the scope of this paper and we leave their discussion for future work. In this work, our goal is to explicitly demonstrate connections in the discrete problem setting, to facilitate further inter-disciplinary work in this area.

\noindent\textbf{Future Work:} 
Finally, we touch upon interesting directions to explore based on the theoretical framework developed in this work.
The first of these is leveraging our derived connections to investigate approximation-based methods from an optimization perspective. Specifically, we propose to work on the characterization of the primal-dual optimization forms of these methods. Since many IRL methods (including adversarial imitation learning) use an optimization perspective, we believe this will not only lead to new algorithmic advances, but will also shed more light on the similarities and differences between our approaches and more recent IRL methods.
Another direction we plan to explore is to use our explicit algorithmic perspectives for practical settings where MCE-IRL is intractable, 
such as problems with very large state spaces, \emph{e.g.} images in activity forecasting. For such situations, our work details how approximation-based methods can be applied in a principled manner when expert data is readily available. We hope to apply our insights to problems such as activity forecasting, social navigation and human preference learning.

\vspace{2mm} 

\noindent {\bf Acknowledgements:} This work was sponsored in part by IARPA (D17PC00340). We thank Dr. Robert A. Miller for his support early in this work.

\bibliography{references}

\begin{appendices}
\noindent{\bf{Main Paper:}} Inverse Reinforcement Learning with Explicit Policy Estimates.

\noindent{\bf{Authors:}} Navyata Sanghvi, Shinnosuke Usami, Mohit Sharma, Joachim Groeger, Kris Kitani.

\noindent{\bf{Published:}} $35^\text{th}$ AAAI Conference on Artificial Intelligence, 2021.

\begin{table}[htbp!]
    \captionsetup{format=myformat}
    \scalebox{0.7}{
        \begin{tabular}{l|l}
        \toprule
        \bf Appendix & \bf Description\\ 
        \midrule
        Appendix \ref{app:A} & Section \ref{sec:theory}: Derivations and Proofs\\
        Appendix \ref{app:B} & Section \ref{sec:ccp-irl}: Optimization-based vs. Approximation-based Methods\\
        Appendix \ref{app:C} & Additional Experiment\\
        Appendix \ref{app:D} & Experimental Details\\
        \bottomrule
        \end{tabular}
    }
\end{table}

\section{Section \ref{sec:theory}: Derivations and Proofs}
\label{app:A}

\subsection{Derivation of the objective gradient (\ref{eqn:likelihood-gradient}).}
\label{app:objective-gradient}
From the objective, soft value and policy form (\ref{eqn:exp-log-likelihood-policy}-\ref{eqn:improved-policy}):
\begin{align}
    g(\bt) &= \mathbb{E}_{P_E(\mathbf{s}, \mathbf{a})}\left[  \textstyle \sum_\tau \log \pi_\bt(a_\tau |s_\tau)  \right]
    \nonumber\\
    &= \textstyle \sum_\tau \mathbb{E}_{P_E(\mathbf{s}_{1:\tau}, \mathbf{a}_{1:\tau})}\left[ \log \pi_\bt(a_\tau |s_\tau)  \right]
    \nonumber\\
    &= \textstyle \sum_\tau \mathbb{E}_{P_E(\mathbf{s}_{1:\tau}, \mathbf{a}_{1:\tau})}\big[ Q^{\tilde{\pi}}_\bt(s_\tau, a_\tau) 
    \nonumber\\
    & \hspace{32mm} - \log \textstyle\sum_{a'} \exp Q^{\tilde{\pi}}_\bt(s_\tau, a')  \big].
    \nonumber
\end{align}
Taking the gradient of this w.r.t. $\bt$:
\begin{align}
    \textstyle \frac{\partial g}{\partial \bt} &= \textstyle \sum_\tau \mathbb{E}_{P_E(\mathbf{s}_{1:\tau}, \mathbf{a}_{1:\tau})}\Big[ \frac{\partial Q^{\tilde{\pi}}_\bt(s_\tau, a_\tau)}{\partial \bt} 
    \nonumber\\
    & \hspace{27mm} - \textstyle\sum_{a'} \frac{ \exp Q^{\tilde{\pi}}_\bt(s_\tau, a') \frac{\partial Q^{\tilde{\pi}}_\bt(s_\tau, a')}{\partial \bt}}{\sum_{a'}  \exp Q^{\tilde{\pi}}_\bt(s_\tau, a') } \Bigg]
    \nonumber\\
    &= \textstyle\sum_\tau \mathbb{E}_{P_E(\mathbf{s}_{1:\tau}, \mathbf{a}_{1:\tau})}\Big[\frac{\partial{Q^{\tilde{\pi}}_\bt(s_\tau, a_\tau)}}{\partial \bt} 
    \nonumber\\
    \label{eqn:derivation-likelihood-gradient}
    & \hspace{27mm} - \textstyle\sum_{a'} \pi_\bt(a'|s_\tau) \frac{\partial{Q^{\tilde{\pi}}_\bt(s_\tau, a')}}{\partial \bt} \Big].
\end{align}
This is the objective gradient \eqref{eqn:likelihood-gradient}. \hfill $\blacksquare$

\subsection{Discounted case: details of objective, soft value, and objective gradient.}
\label{app:general-form-discounted}
The general forms we described in (\ref{eqn:exp-log-likelihood-policy}, \ref{eqn:soft-value}, \ref{eqn:likelihood-gradient}) consider no discounting. In case of discounting by a factor $\gamma$, simple modifications apply to the objective, soft value and objective gradient.
\begin{align}
    \label{eqn:exp-log-likelihood-policy-discount}
    & g(\bt) \;\;\;\;\;\;\;=\; \mathbb{E}_{P_E(\mathbf{s}, \mathbf{a})}\left[ \textstyle\sum_\tau \gamma^\tau \log \pi_{\bt}(a_\tau | s_\tau) \right].
    \\
    & Q^{\tilde{\pi}}_\bt(s,a) \;=\; r(s,a,\bt) + \gamma \mathbb{E}_{s' \sim p', a' \sim \tilde{\pi}} \big[
    \nonumber \\
    \label{eqn:soft-value-discount}
    & \hspace{35mm} Q^{\tilde{\pi}}_\bt(s',a') - \log \tilde{\pi}(a'|s') \big].
    \\
    & \textstyle\frac{\partial{g}}{\partial{\bt}} \;\;\;\;\;\;\;\;\;\;=\; \textstyle\sum_\tau \gamma^\tau \mathbb{E}_{P_E(\mathbf{s}_{1:\tau}, \mathbf{a}_{1:\tau})}\Big[\frac{\partial{Q^{\tilde{\pi}}_\bt(s_\tau, a_\tau)}}{\partial \bt} 
    \nonumber \\
    \label{eqn:likelihood-gradient-discount}
    & \hspace{30mm} - \textstyle\sum_{a'} \pi_\bt(a'|s_\tau) \frac{\partial{Q^{\tilde{\pi}}_\bt(s_\tau, a')}}{\partial \bt} \Big].
\end{align}

\subsection{Proof that NFXP value function {$Q^*_\bt$} is the optimal ``soft'' value, and that NFXP policy {$\pi_\bt$} is optimal in the soft value sense.}
\label{app:nfxp-proof-pitilde}

{\bf Optimal ``soft'' value:} First, let us talk about the optimal ``soft'' value \cite{haarnoja2018soft}. 
For this proof, we use the discounted definition of the soft value \eqref{eqn:soft-value-discount} from Appendix \ref{app:general-form-discounted}, since the NFXP work \cite{rust_theory} assumes discounting by a factor $\gamma$.
We observe that, under the optimal policy $\pi_\bt^*$ (in the soft value sense), the soft value function (\ref{eqn:soft-value-discount}) satisfies:
\begin{align}
    Q^{\pi^*_\bt}_\bt(s,a) &= r(s,a,\bt) + \gamma \mathbb{E}_{s' \sim p', a' \sim \tilde{\pi}} \Big[
    \nonumber\\
    \label{eqn:nfxp-optimal-value}
    & \hspace{15mm} Q^{\pi^*_\bt}_\bt(s',a') - \log \pi^*_\bt(a'|s') \Big].
\end{align}
From \cite{haarnoja2018soft}, the optimal policy $\pi^*_\bt$ satisfies: 
\begin{align*}
     \pi^*_\bt(a|s) = \text{softmax } Q^{\pi^*_\bt}_{\bt}(s,a)
\end{align*}


\noindent From this and \eqref{eqn:nfxp-optimal-value}: 
\begin{align}
    Q^{\pi^*_\bt}_{\bt}(s,a) &= r(s,a,\bt) 
    \nonumber\\
    \label{eqn:optimal-soft-value}
    & \hspace{5mm} + \gamma \mathbb{E}_{s' \sim p'}\left[ \log \textstyle\sum_{a'}\exp Q^{\pi^*_\bt}_{\bt}(s', a') \right]
\end{align}

\noindent {\bf Forward problem (Sec. \ref{sec:prelims}):} 
Since the forward problem solution $Q^*_\bt$ is the {\bf unique} fixed point of (\ref{eqn:nfxp-contraction}), thus, (\ref{eqn:nfxp-contraction}, \ref{eqn:optimal-soft-value}) imply that $Q^*_\bt = Q^{\pi^*_\bt}_\bt$. Thus,  $Q^*_\bt$ is equivalent to the optimal ``soft'' value. \hfill $\blacksquare$

\noindent {\bf NFXP policy $\pi_\bt$ is optimal in the soft value sense:} We have shown $Q^*_\bt  = Q^{\pi^*_\bt}_\bt$. Thus, the policy that attains the soft value $Q^*_\bt$ is $\pi^*_\bt$. From this, (\ref{eqn:nfxp-policy}) and \cite{haarnoja2018soft}, we can conclude that $\pi_\bt = \pi^*_\bt$. Therefore, $\pi_\bt$ is the optimal policy in the soft value sense. \hfill $\blacksquare$


\subsection{Derivation of NFXP gradient (\ref{eqn:nfxp-gradient}).}
\label{app:nfxp-gradient}
Since the NFXP work \cite{rust_theory} considers discounting, we consider maximization of the following objective of expected log likelihood of trajectories generated using $\pi_\bt$ \eqref{eqn:nfxp-policy}:
\begin{align}
    & \mathbb{E}_{P_E(\mathbf{s}, \mathbf{a})}\left[ \textstyle\sum_\tau \gamma^\tau \log \pi_{\bt}(a_\tau | s_\tau) \right]
    \nonumber\\
    & = \textstyle \sum_\tau \gamma^\tau \mathbb{E}_{P_E(\mathbf{s}_{1:\tau}, \mathbf{a}_{1:\tau})}\Big[ Q^*_\bt(s_\tau, a_\tau) 
    \nonumber \\
    & \hspace{35mm} - \log \textstyle\sum_{a'} \exp Q^*_\bt(s_\tau, a')  \Big].
    \nonumber
\end{align}
Taking the gradient of this wrt $\bt$, we get:
\begin{align}
    \textstyle\sum_\tau \gamma^\tau \mathbb{E}_{P_E(\mathbf{s}_{1:\tau}, \mathbf{a}_{1:\tau})}
    & \Big[\textstyle\frac{\partial{Q^*_\bt(s_\tau, a_\tau)}}{\partial \bt} 
    \nonumber \\
    \label{eqn:proof-nfxp-gradient}
    & - \textstyle\sum_{a'} \pi_\bt(a'|s_\tau) \frac{\partial{Q^*_\bt(s_\tau, a')}}{\partial \bt} \Big]
\end{align}

\noindent {\bf Derivative of $Q^*_\bt$ w.r.t. $\bt$:} As detailed in \eqref{eqn:nfxp-contraction}, $Q^*_\bt$ is the unique fixed point of the contraction mapping $\Lambda_\bt$:
\begin{align}
    \Lambda_\bt(Q)(s,a) = \;& r(s,a,\bt) 
    \nonumber\\
    \label{eqn:proof-nfxp-contraction}
    & + \gamma \mathbb{E}_{s' \sim p'}\left[ \log \textstyle\sum_{a'}\exp Q(s', a') \right]
\end{align}

\noindent The partial derivative $\frac{\partial{Q^*_\bt}}{\partial \bt}$ is given in \cite{rust_theory} in terms of derivatives of $\Lambda_\bt$:
\begin{align}
    \label{eqn:proof-dq-dtheta}
    \textstyle \frac{\partial{Q^*_\bt}}{\partial \bt} = \left[I - \Lambda'_\bt(Q)\right]^{-1} \left. \left[\frac{\partial \Lambda_\bt(Q)}{\partial \bt}\right] \right|_{Q=Q^*_\bt}
\end{align}

\noindent {\bf Derivative of $\Lambda_\bt$ w.r.t. $Q$:} From \eqref{eqn:proof-nfxp-contraction}, we can derive $\Lambda'_\bt(Q)$ (\emph{i.e.}, the derivative of $\Lambda_\bt$ w.r.t. $Q$) as an $|\mathcal{S} \times \mathcal{A}| \times |\mathcal{S} \times \mathcal{A}|$ matrix. We denote the element in the $i^{\text{th}}$ row and $j^{\text{th}}$ column as: $\left(\Lambda'_\bt(Q)\right)_{ij}$. This element corresponds to the $i^\text{th}$ state-action pair $(s_i,a_i)$ as input to the operator and the $j^{\text{th}}$ state-action pair $(s_j,a_j)$ as output. From \eqref{eqn:proof-nfxp-contraction}:
{\scriptsize
{\begin{align}
    \left(\Lambda'_\bt(Q)\right)_{ij} &= \frac{\partial \left(r(s_i, a_i, \bt) + \gamma \mathbb{E}_{s' \sim T(s'|s_i,a_i)}\left[ \log \textstyle\sum_{a'}\exp Q(s', a') \right] \right)}{\partial Q(s_j,a_j)} 
    \nonumber\\
    &= \gamma T(s_j | s_i, a_i) \frac{\exp \; Q(s_j, a_j)}{\textstyle\sum_{a'}\exp \; Q(s_j, a')} 
    \nonumber\\
    &= \gamma T(s_j | s_i, a_i) \;\text{softmax} \; Q(s_j,a_j)
    \nonumber \\
    \label{eqn:proof-dl-dq}
    &= \gamma T(s_j | s_i, a_i) \;\pi_\bt(a_j|s_j), \hspace{7mm} \text{when } Q = Q^*_\bt \;\; \eqref{eqn:nfxp-policy}.
\end{align}
}}%

\noindent {\bf Derivative of $\Lambda_\bt$ w.r.t. $\bt$:} Let us denote the number of parameters in $\bt$ as $n$. From \eqref{eqn:proof-nfxp-contraction}, we can derive $\frac{\partial \Lambda_\bt(Q)}{\partial \bt}$ (\emph{i.e.}, the derivative of $\Lambda_\bt$ w.r.t. $\bt$) as an $|\mathcal{S} \times \mathcal{A}| \times n$ matrix. The element in the $i^{\text{th}}$ row and $j^{\text{th}}$ column corresponds to the $i^\text{th}$ state-action pair $(s_i,a_i)$ as input and the $j^\text{th}$ parameter $\theta_j$ as output. From \eqref{eqn:proof-nfxp-contraction}:
\begin{align}
    \label{eqn:proof-dl-dtheta}
    \textstyle \left(\frac{\partial \Lambda_\bt(Q)}{\partial \bt}\right)_{ij} = \frac{\partial r(s_i,a_i,\bt)}{\partial \theta_j}
\end{align}

\noindent {\bf NFXP gradient:} Plugging the derived (\ref{eqn:proof-dl-dq}, \ref{eqn:proof-dl-dtheta}) into \eqref{eqn:proof-dq-dtheta}, we have:
\begin{align}
    \textstyle \frac{\partial{Q^*_\bt}(s,a)}{\partial \bt} =\; & \mathbb{E}_{P_\bt(\mathbf{s}_{>0}, \mathbf{a}_{>0})}\Big[
    \nonumber \\
    & \textstyle \sum_{t \geq 0} \gamma^t \frac{\partial r(s_t,a_t,\bt)}{\partial \bt} \Big| (s_0,a_0)=(s,a) \Big]
\end{align}

\noindent Finally, putting this derived value gradient into \eqref{eqn:proof-nfxp-gradient}, the gradient of the NFXP objective is:
{{
\begin{align}
    & \textstyle \sum_\tau \gamma^\tau \mathbb{E}_{P_E(\mathbf{s}_{1:\tau}, \mathbf{a}_{1:\tau-1})}\Big[
    \nonumber\\
    & \hspace{13mm} \mathbb{E}_{\pi_E(a_\tau|s_\tau), P_\bt(\mathbf{s}_{>\tau}, \mathbf{a}_{>\tau})} \left[\textstyle\sum_{t \geq \tau} \gamma^{t-\tau} \frac{\partial{r(s_t, a_t, \bt)}}{\partial \bt}\right] 
    \nonumber\\
    & \hspace{10mm} -  \left.\mathbb{E}_{\pi_\bt(a_\tau|s_\tau), P_\bt(\mathbf{s}_{>\tau}, \mathbf{a}_{>\tau})} \left[\textstyle \sum_{t \geq \tau} \gamma^{t-\tau} \frac{\partial{r(s_t, a_t, \bt)}}{\partial \bt}\right] \right]
    \nonumber\\
    =& \; \textstyle \sum_\tau  \mathbb{E}_{P_E(\mathbf{s}_{1:\tau}, \mathbf{a}_{1:\tau-1})}\Big[
    \nonumber\\
    & \hspace{13mm}\mathbb{E}_{\pi_E(a_\tau|s_\tau), P_\bt(\mathbf{s}_{>\tau}, \mathbf{a}_{>\tau})} \left[\textstyle\sum_{t \geq 0} \gamma^{t} \frac{\partial{r(s_t, a_t, \bt)}}{\partial \bt}\right] 
    \nonumber\\
    & \hspace{10mm} -  \left.\mathbb{E}_{\pi_\bt(a_\tau|s_\tau), P_\bt(\mathbf{s}_{>\tau}, \mathbf{a}_{>\tau})} \left[\textstyle \sum_{t \geq 0} \gamma^{t} \frac{\partial{r(s_t, a_t, \bt)}}{\partial \bt}\right] \right]
    \nonumber\\
    =& \; \mathbb{E}_{P_E(\mathbf{s}, \mathbf{a})}\left[\textstyle \sum_t \gamma^t \frac{\partial{r(s_t, a_t, \bt)}}{\partial \bt} \right] 
    \nonumber\\
    \label{eqn:proof-nfxp-final-gradient}
    & \hspace{5mm} - \mathbb{E}_{P_\bt(\mathbf{s}, \mathbf{a})}\left[\textstyle \sum_t \gamma^t  \frac{\partial{r(s_t, a_t, \bt)}}{\partial \bt} \right].
\end{align}
}}%

\noindent Since we compare NFXP with MCE-IRL, it is only fair if we consider the \emph{same} discounting for NFXP that we consider for MCE-IRL. Therefore, using discount factor $\gamma = 1$, the gradient we derived in \eqref{eqn:proof-nfxp-final-gradient} is the NFXP gradient \eqref{eqn:nfxp-gradient}. \hfill $\blacksquare$



\subsection{Equivalence of objective gradient (\ref{eqn:likelihood-gradient}) with NFXP and MCE-IRL gradients (\ref{eqn:nfxp-gradient}, \ref{eqn:maxent-gradient}).}
\label{app:nfxp-equi-gradient}    

We have shown that we can specify $\tilde{\pi} = \pi_\bt$ in the general policy form \eqref{eqn:improved-policy}, for both NFXP and MCE-IRL. Using this specific $\tilde{\pi}$, we can prove that the objective gradients resulting from \eqref{eqn:likelihood-gradient} are the same as those of NFXP and MCE-IRL.

\noindent {\bf NFXP:} The NFXP work \cite{rust_theory} assumes discounting by a factor $\gamma$. Thus, we start our proof using the objective, value and gradient in Appendix \ref{app:general-form-discounted}. Using $\tilde{\pi} = \pi_\bt$ in Eqns. (\ref{eqn:exp-log-likelihood-policy-discount}-\ref{eqn:likelihood-gradient-discount}):
{{
\begin{align}
    \textstyle \frac{\partial{g}}{\partial{\bt}} &= \textstyle \sum_\tau \gamma^\tau \mathbb{E}_{P_E(\mathbf{s}_{1:\tau}, \mathbf{a}_{1:\tau})}\Big[\frac{\partial{Q^{\pi_\bt}_\bt(s_\tau, a_\tau)}}{\partial \bt}
    \nonumber\\
    & \hspace{10mm} - \textstyle\sum_{a'} \pi_\bt(a'|s_\tau) \frac{\partial{Q^{\pi_\bt}_\bt(s_\tau, a')}}{\partial \bt} \Big]
    \nonumber\\
    &= \textstyle \sum_\tau \gamma^\tau  \mathbb{E}_{P_E(\mathbf{s}_{1:\tau}, \mathbf{a}_{1:\tau-1})}\Big[
    \sum_{a'} \pi_E(a'|s_\tau) \frac{\partial{Q^{\pi_\bt}_\bt(s_\tau, a')}}{\partial \bt}  
    \nonumber\\
    & \hspace{32mm} - \textstyle\sum_{a'} \pi_\bt(a'|s_\tau) \frac{\partial{Q^{\pi_\bt}_\bt(s_\tau, a')}}{\partial \bt} \Big]
    \nonumber\\
    &= \textstyle \sum_\tau \gamma^\tau \mathbb{E}_{P_E(\mathbf{s}_{1:\tau}, \mathbf{a}_{1:\tau-1})}\Big[
    \nonumber\\
    & \hspace{13mm} \mathbb{E}_{\pi_E(a_\tau|s_\tau), P_\bt(\mathbf{s}_{>\tau}, \mathbf{a}_{>\tau})} \left[\textstyle\sum_{t \geq \tau} \gamma^{t-\tau} \frac{\partial{r(s_t, a_t, \bt)}}{\partial \bt}\right] 
    \nonumber\\
    & \hspace{10mm} -  \left.\mathbb{E}_{\pi_\bt(a_\tau|s_\tau), P_\bt(\mathbf{s}_{>\tau}, \mathbf{a}_{>\tau})} \left[\textstyle \sum_{t \geq \tau} \gamma^{t-\tau} \frac{\partial{r(s_t, a_t, \bt)}}{\partial \bt}\right] \right]
    \nonumber\\
    &= \textstyle \sum_\tau  \mathbb{E}_{P_E(\mathbf{s}_{1:\tau}, \mathbf{a}_{1:\tau-1})}\Big[
    \nonumber\\
    & \hspace{13mm} \mathbb{E}_{\pi_E(a_\tau|s_\tau), P_\bt(\mathbf{s}_{>\tau}, \mathbf{a}_{>\tau})} \left[\textstyle\sum_{t \geq 0} \gamma^{t} \frac{\partial{r(s_t, a_t, \bt)}}{\partial \bt}\right] 
    \nonumber\\
    & \hspace{10mm} -  \left.\mathbb{E}_{\pi_\bt(a_\tau|s_\tau), P_\bt(\mathbf{s}_{>\tau}, \mathbf{a}_{>\tau})} \left[\textstyle \sum_{t \geq 0} \gamma^{t} \frac{\partial{r(s_t, a_t, \bt)}}{\partial \bt}\right] \right]
    \nonumber\\
    &= \mathbb{E}_{P_E(\mathbf{s}, \mathbf{a})}\left[\textstyle \sum_t \gamma^t \frac{\partial{r(s_t, a_t, \bt)}}{\partial \bt} \right] 
    \nonumber\\
    \label{eqn:proof-nfxp-obje-gradient}
    & \hspace{10mm} - \mathbb{E}_{P_\bt(\mathbf{s}, \mathbf{a})}\left[\textstyle \sum_t \gamma^t  \frac{\partial{r(s_t, a_t, \bt)}}{\partial \bt} \right].
\end{align}
}}%
We can only compare the methods NFXP and MCE-IRL under the \emph{same} assumptions of discounting, i.e., $\gamma=1$. Using this discount factor: \;\; 1) the forms (\ref{eqn:exp-log-likelihood-policy-discount}-\ref{eqn:likelihood-gradient-discount}) are the same as (\ref{eqn:exp-log-likelihood-policy}-\ref{eqn:likelihood-gradient}), and, \;\; 2) the resulting objective gradient \eqref{eqn:proof-nfxp-obje-gradient} is equivalent to the NFXP gradient \eqref{eqn:nfxp-gradient}. \hfill $\blacksquare$

\noindent {\bf MCE-IRL:} Using $\tilde{\pi} = \pi_\bt$, and linear reward $r(s,a,\bt) = \bt^T\mathbf{f}(s,a)$ in Eqns. (\ref{eqn:soft-value}-\ref{eqn:likelihood-gradient}):
{{
\begin{align}
    \textstyle \frac{\partial{g}}{\partial{\bt}} &= \textstyle \sum_\tau \mathbb{E}_{P_E(\mathbf{s}_{1:\tau}, \mathbf{a}_{1:\tau})}\Big[\frac{\partial{Q^{\pi_\bt}_\bt(s_\tau, a_\tau)}}{\partial \bt} 
    \nonumber\\
    & \hspace{10mm} - \textstyle\sum_{a'} \pi_\bt(a'|s_\tau) \frac{\partial{Q^{\pi_\bt}_\bt(s_\tau, a')}}{\partial \bt} \Big]
    \nonumber\\
    &= \textstyle \sum_\tau \mathbb{E}_{P_E(\mathbf{s}_{1:\tau}, \mathbf{a}_{1:\tau-1})}\Big[
    \sum_{a'} \pi_E(a'|s_\tau) \frac{\partial{Q^{\pi_\bt}_\bt(s_\tau, a')}}{\partial \bt} 
    \nonumber\\
    & \hspace{28mm} - \textstyle\sum_{a'} \pi_\bt(a'|s_\tau) \frac{\partial{Q^{\pi_\bt}_\bt(s_\tau, a')}}{\partial \bt} \Big]
    \nonumber\\
    &= \textstyle \sum_\tau \mathbb{E}_{P_E(\mathbf{s}_{1:\tau}, \mathbf{a}_{1:\tau-1})}\Big[
    \nonumber\\
    & \hspace{18mm} \mathbb{E}_{\pi_E(a_\tau|s_\tau), P_\bt(\mathbf{s}_{>\tau}, \mathbf{a}_{>\tau})} \left[\textstyle\sum_{t \geq \tau} \frac{\partial{r(s_t, a_t, \bt)}}{\partial \bt}\right] 
    \nonumber\\
    & \hspace{15mm} -  \left.\mathbb{E}_{\pi_\bt(a_\tau|s_\tau), P_\bt(\mathbf{s}_{>\tau}, \mathbf{a}_{>\tau})} \left[\textstyle \sum_{t \geq \tau} \frac{\partial{r(s_t, a_t, \bt)}}{\partial \bt}\right] \right]
    \nonumber\\
    &= \textstyle \sum_\tau \mathbb{E}_{P_E(\mathbf{s}_{1:\tau}, \mathbf{a}_{1:\tau-1})}\Big[
    \nonumber\\
    & \hspace{18mm} \mathbb{E}_{\pi_E(a_\tau|s_\tau), P_\bt(\mathbf{s}_{>\tau}, \mathbf{a}_{>\tau})} \left[\textstyle\sum_{t \geq 0} \frac{\partial{r(s_t, a_t, \bt)}}{\partial \bt}\right] 
    \nonumber\\
    & \hspace{15mm} -  \left.\mathbb{E}_{\pi_\bt(a_\tau|s_\tau), P_\bt(\mathbf{s}_{>\tau}, \mathbf{a}_{>\tau})} \left[\textstyle \sum_{t \geq 0} \frac{\partial{r(s_t, a_t, \bt)}}{\partial \bt}\right] \right]
    \nonumber\\
    &= \mathbb{E}_{P_E(\mathbf{s}, \mathbf{a})}\left[\textstyle \sum_t \mathbf{f}(s_t, a_t) \right] 
    \nonumber\\
    \label{eqn:proof-maxent-grad}
    & \hspace{10mm} - \mathbb{E}_{P_\bt(\mathbf{s}, \mathbf{a})}\left[\textstyle \sum_t \mathbf{f}(s_t, a_t) \right].
\end{align}
}}%
The resulting objective gradient \eqref{eqn:proof-maxent-grad} is equivalent to the MCE-IRL gradient \eqref{eqn:maxent-gradient}. \hfill $\blacksquare$

\section{Section \ref{sec:ccp-irl}: Optimization-based vs. Approximation-based Methods}
\label{app:B}

We discuss {\bf matrix representations} of computations required in each algorithm (Alg. \ref{algo:optimization} and \ref{algo:approx}), focusing on gradient steps. Through matrix representations, we discuss what can be pre-computed and what must be computed anew in each step of following the gradient. We demonstrate the high computational {\bf burden} of the gradient step in optimization-based methods when compared to approximation-based methods.

\subsection{Notation}

\begin{itemize}[leftmargin=*, topsep=0pt, partopsep=0pt]
    \item {\bf State distributions and transitions:}
    \\$\mathbf{P}_s$ represents the $1 \times |\mathcal{S}|$ vector of initial state probabilities.
    \\ $\mathbf{T}_{sas'}$ represents the $|\mathcal{S}| \times |\mathcal{A}| \times |\mathcal{S}|$ tensor of transition probabilities $T(s'|s,a)$.
    \\ $\mathbf{T}^{\pi}_{ss'}$ represents the $|\mathcal{S}| \times |\mathcal{S}|$ matrix of transition probabilities $P^\pi(s'|s) = \sum_{a} \pi(a|s) T(s'|s,a)$.
    \item {\bf Occupancy measures:} 
    \\$\mathbf{O}_{s'}^{\pi}$ represents the $1 \times |\mathcal{S}|$ vector of occupancy measures $\text{Occ}^\pi(s')$.
    \\$\mathbf{O}_{sas'}^{\pi}$ represents the $|\mathcal{S}| \times |\mathcal{A}| \times |\mathcal{S}|$ tensor of occupancy measures $\text{Occ}^\pi(s'|s,a)$.
    \item {\bf Rewards and entropy:}
    \\$\mathbf{R}^\bt_{sa}$ represents the $|\mathcal{S}| \times |\mathcal{A}|$ matrix of rewards $r(s,a,\bt)$.
    \\$\mathbf{L}^\pi_{s}$ represents the $|\mathcal{S}| \times 1$ vector of expected reward and entropy $L^\pi(s) = \sum_a \pi(a|s) \left(r(s,a,\bt) - \log \pi(a|s) \right)$.
    \item {\bf Reward derivatives:}
    \\$\mathbf{D}_{sa\theta}$ represents the $|\mathcal{S}| \times |\mathcal{A}| \times n$ tensor of reward function derivatives $\frac{\partial{r(s, a, \bt)}}{\partial \bt}$.
    \\$\mathbf{E}^\pi_{s\theta}$ represents the $|\mathcal{S}| \times n$ matrix of expected reward derivatives $E^\pi(s,i) = \sum_a \pi(a|s) \frac{\partial{r(s, a, \bt)}}{\partial \theta_i}$.
    $n=$ number of parameters $\bt$.
    \item {\bf Soft value derivatives:}
    \\$\mathbf{EQ}^{\pi}_{s\theta}$ represents the $|\mathcal{S}| \times n$ matrix of expected value derivatives $EQ^\pi(s,i) = \sum_a \pi(a|s) \frac{\partial{Q^{\tilde{\pi}}_\bt(s, a)}}{\partial \theta_i}$
\end{itemize}

\subsection{Optimization-based Methods}
\label{app:sub:optim}
In Table \ref{tab:optim}, we summarize details of computation required in Alg. \ref{algo:optimization}. Dynamic programming is abbreviated as {\bf DP}.

\begin{table*}[htbp!]
    \centering
    \captionsetup{format=myformat}
    \scalebox{0.9}{
        \begin{tabular}{l||ll|l|l|l}
        \toprule
        \bf When? & \bf Description & \bf Computation & \bf Lines & \bf Requirement & \bf Frequency \\
        \midrule
        \midrule
        \bf Pre-computed & Occ. under expert policy $\hat{\pi}_E$:\;\;\; & $\mathbf{O}^{\hat{\pi}_E}_{s'} = \mathbf{P}_s\left(\mathbf{I} - \gamma \mathbf{T}^{\hat{\pi}_E}_{ss'}\right)^{-1}$ & Line 2 & {\bf DP} / Matrix inverse & Once. \\
        \midrule
        \multirow{4}{*}{\bf Within loop} & 1. Optimal soft value and policy: & $Q^*_\bt$ \eqref{eqn:nfxp-contraction},\;\;\; $\pi_\bt$ \eqref{eqn:nfxp-policy}.\;\;\;\;\; & Lines 4-5 & {\bf DP} / Value Iteration &   \\
        & 2. Occ. under optimal policy $\pi_\bt$:\;\;\; &
        $\mathbf{O}^{\pi_\bt}_{s'} = \mathbf{P}_s\left(\mathbf{I} - \gamma \mathbf{T}^{\pi_\bt}_{ss'}\right)^{-1}$ & Line 6 & {\bf DP} / Matrix inverse & Once per \\
        & 3. Exp. reward deriv. counts under $\pi_\bt$: & $\boldsymbol{\mu}^{\pi_\bt} \;= \mathbf{O}_{s'}^{\pi_\bt} \mathbf{E}^{\pi_\bt}_{s\theta}$ & Line 6 & Matrix multiplication & $\bt$ update. \\
        & 4. Objective gradient: & $\frac{\partial g}{\partial \bt} \;\; = \mathbf{O}_{s'}^{\hat{\pi}_E} \mathbf{E}^{\hat{\pi}_E}_{s\theta} - \boldsymbol{\mu}^{\pi_\bt}$ & Line 7 & Matrix multiplication & \\
        \bottomrule
        \end{tabular}
    }
    \caption{Optimization-based methods: Computational requirements.}
    \label{tab:optim}
\end{table*}

\subsection{Approximation-based Methods}
\label{app:sub:approx}
In Table \ref{tab:approx}, we summarize details of computation required in Alg. \ref{algo:approx}. Dynamic programming is abbreviated as {\bf DP}.

\begin{table*}[htbp!]
    \centering
    \captionsetup{format=myformat}
    \scalebox{0.9}{
        \begin{tabular}{l||ll|l|l|l}
        \toprule
        \bf When? & \bf Description & \bf Computation & \bf Lines & \bf Requirement & \bf Frequency \\
        \midrule
        \midrule
        \bf Pre-computed & Occ. under expert policy $\hat{\pi}_E$:\; & $\mathbf{O}^{\hat{\pi}_E}_{s'} = \mathbf{P}_s\left(\mathbf{I} - \gamma \mathbf{T}^{\hat{\pi}_E}_{ss'}\right)^{-1}$ & Line 3 & {\bf DP} / Matrix inverse & Once. \\
        \midrule
        \bf Within & \multirow{2}{*}{Occ. under policy $\tilde{\pi}$:\;} &
        \multirow{2}{*}{$\mathbf{O}^{\tilde{\pi}}_{sas'} = \mathbf{T}_{sas'}\left(\mathbf{I} - \gamma \mathbf{T}^{\tilde{\pi}}_{ss'}\right)^{-1}$} & \multirow{2}{*}{Line 5} & \multirow{2}{*}{{\bf DP} / Matrix inverse} & Once per \\
        \bf outer loop & & & & &  $\tilde{\pi}$ update. \\
        \midrule
        \multirow{4}{*}{\shortstack[l]{\bf Within \\ \bf inner loop}}& 1. Soft value and policy: & $Q^{\tilde{\pi}}_\bt \;\;\; = \mathbf{R}^\bt_{sa} \;\; + \mathbf{O}^{\tilde{\pi}}_{sas'} \mathbf{L}^{\tilde{\pi}}_{s}$, \;\;$\pi_\bt$ \eqref{eqn:improved-policy} & Lines 7-8 & Matrix multiplication & \multirow{4}{*}{\shortstack[l]{ Once per \\ $\bt$ update.}} \\
        & 2. Value derivatives: & $\frac{\partial Q^{\tilde{\pi}_\bt}}{\partial \bt} = \mathbf{D}_{sa\theta} + \mathbf{O}_{sas'}^{\tilde{\pi}} \mathbf{E}^{\tilde{\pi}}_{s\theta}$ & Line 9 & Matrix multiplication & \\
        & 3. Objective gradient: & $\frac{\partial g}{\partial \bt} \;\;\;\; = \mathbf{O}_{s'}^{\hat{\pi}_E} \left(
        \mathbf{EQ}^{\hat{\pi}_E}_{s\theta} -
        \mathbf{EQ}^{\pi_\bt}_{s\theta} \right)$ & Line 10 & Matrix multiplication & \\
        \bottomrule
        \end{tabular}
    }
    \caption{Approximation-based methods: Computational requirements.}
    \label{tab:approx}
    \vspace{-4mm}
\end{table*}

\subsection{Comparison}
Using Appendices \ref{app:sub:optim} and \ref{app:sub:approx}, we can discuss how optimization-based methods have a {\bf much higher} computational burden than approximation-based methods:
\begin{enumerate}[leftmargin=*, topsep=0pt, partopsep=0pt]
    \item {\bf Pre-computation:} From the ``Pre-computed'' row in Tables \ref{tab:optim} and \ref{tab:approx}, the pre-computation requirements are the {\bf same} for both optimization-based and approximation-based methods.
    \item {\bf $\bt$ update loop:} From Table \ref{tab:optim}: ``Within loop'' and Table \ref{tab:approx}: ``Within inner loop'', we see that optimization-based methods require very expensive value iteration and matrix inverse at {\bf every} gradient step. On the other hand, approximation-based methods require {\bf no} dynamic programming. The computational requirement of ``Matrix multiplication'' is much cheaper than that of dynamic programming, and is of the order of the size of the environment. Thus, the computational requirement of the $\bt$ update is {\bf much larger} for optimization-based methods than for approximation-based methods.
    \item {\bf Policy update loop:} From Table \ref{tab:approx}: ``Within outer loop'', we see that approximation-based methods have an outer loop of $\tilde{\pi}$ update, which requires computing a matrix inverse. Optimization-based methods have no outer loop. Although this step requires dynamic programming, the number of policy updates is {\bf many magnitudes fewer} (order of 10) than the number of $\bt$ updates (order of 1000's). Therefore, the number of dynamic programming steps that optimization-based methods require is {\bf many magnitudes more} than the number required by approximation-based methods.
\end{enumerate}

\subsection{Example Time Breakdown}
In Table \ref{tab:optim_time_breakdown} and \ref{tab:approx_time_breakdown}, we show an example time breakdown on Mountain Car with $100^{2}$ discretized states and Objectworld with $64^{2}$ grid and 4 colors. Each number represents the averaged time for computing the corresponding step once.
\begin{table*}[htbp!]
    \centering
    \captionsetup{format=myformat}
    \scalebox{0.9}{
        \begin{tabular}{l||l|l|rr}
        \toprule
        \bf When? & \bf Computation & \bf Requirement & \bf Mountain Car & \bf Objectworld \\
        \midrule
        \midrule
        \bf Pre-computed & Occ. under expert policy $\hat{\pi}_E$ \;\;\; & {\bf DP} / Matrix inverse & \bf 8.97 & \bf 1.72 \\
        \midrule
        \multirow{5}{*}{\bf Within loop} & 1. Optimal soft value and policy & {\bf DP} / Value Iteration & 105.75 & 2.91 \\
        & 2. Occ. under optimal policy $\pi_\bt$\;\;\; &{\bf DP} / Matrix inverse 
        & {9.07} & {2.13} \\
        & 3. Exp. reward deriv. counts under $\pi_\bt$ & Matrix multiplication &  0.01 & 0.00   \\
        & 4. Objective gradient & Matrix multiplication & 0.03 & 0.10 \\
        \cline{2-5}\\[-2.5mm]
        & \multicolumn{2}{r}{{\bf Inner Loop  Per-Step Totals:} } & \bf 114.86 & \bf 5.14 \\
        \bottomrule
        \end{tabular}
    }
    \caption{Optimization-based methods: Example time breakdown (in seconds).}
    \label{tab:optim_time_breakdown}
    \vspace{-2mm}
\end{table*}

\begin{table*}[htbp!]
    \centering
    \captionsetup{format=myformat}
    \scalebox{0.9}{
        \begin{tabular}{l||l|l|rr}
        \toprule
        \bf When? & \bf Computation & \bf Requirement & \bf Mountain Car & \bf Objectworld \\
        \midrule
        \midrule
        \bf Pre-computed & Occ. under expert policy $\hat{\pi}_E$\; & {\bf DP} / Matrix inverse & \bf 13.50 & \bf 2.69 \\
        \midrule
        \bf Within & \multirow{2}{*}{Occ. under policy $\tilde{\pi}$\;} &
        \multirow{2}{*}{{\bf DP} / Matrix inverse}  & \multirow{2}{*}{\bf 24.39} & \multirow{2}{*}{\bf 2.11} \\
        \bf outer loop & & & &  \\
        \midrule
        \multirow{4}{*}{\shortstack[l]{\bf Within \\ \bf inner loop}}& 1. Soft value and policy & Matrix multiplication & 0.11 & 0.12 \\
        & 2. Value derivatives & Matrix multiplication & 1.36 & 0.01\\
        & 3. Objective gradient & Matrix multiplication & 0.37 & 0.01\\
        \cline{2-5}\\[-2.5mm]
        & \multicolumn{2}{r}{{\bf Inner Loop Per-Step Totals:} } & \bf 1.84 & \bf 0.14 \\
        \bottomrule
        \end{tabular}
    }
    \caption{Approximation-based methods: Example time breakdown (in seconds).}
    \label{tab:approx_time_breakdown}
    \vspace{-5mm}
\end{table*}

From Table \ref{tab:optim_time_breakdown}, in optimization-based methods, the computation of the optimal soft value and policy within the $\bt$ update loop requires a large amount of time ({\bf DP} steps). This leads to a large computational burden because the optimal soft value and policy computation has to be done every time $\bt$ is updated (order of 1000 times). On the other hand, from Table \ref{tab:approx_time_breakdown}, in approximation-based methods, pre-computation of the occupancy under policy $\hat{\pi}_E$ and policy $\tilde{\pi}$ takes the most time ({\bf DP} steps), but still not as much as the inner loop in optimization-based methods. Moreover, the inner loop computation takes much lesser time because there are no {\bf DP} steps in it (\emph{e.g.,} 114.86 s vs. 1.84 s for MountainCar). Since occupancy under $\hat{\pi}_E$  has to be computed only once and the occupancy under $\tilde{\pi}$ has to be computed only once per outer loop (order of 10 times), this is much more computationally efficient than optimization-based methods.

\section{Additional Experiment}
\label{app:C}

\begin{figure*}[h]
\centering
 \begin{tabular}{ccc}
    \MSHangBox{\includegraphics[width=0.2\textwidth]{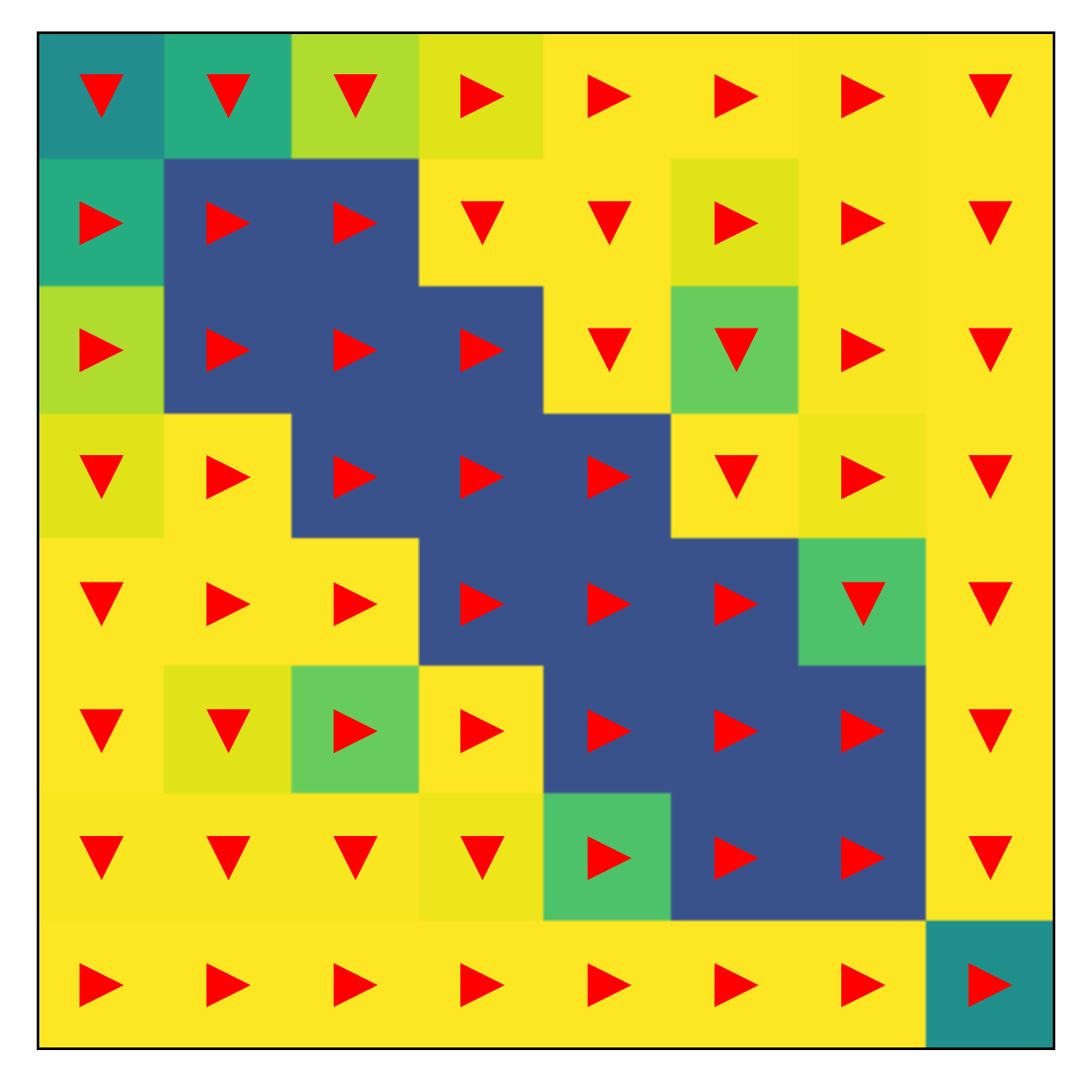}}&
    \MSHangBox{\includegraphics[width=0.25\textwidth]{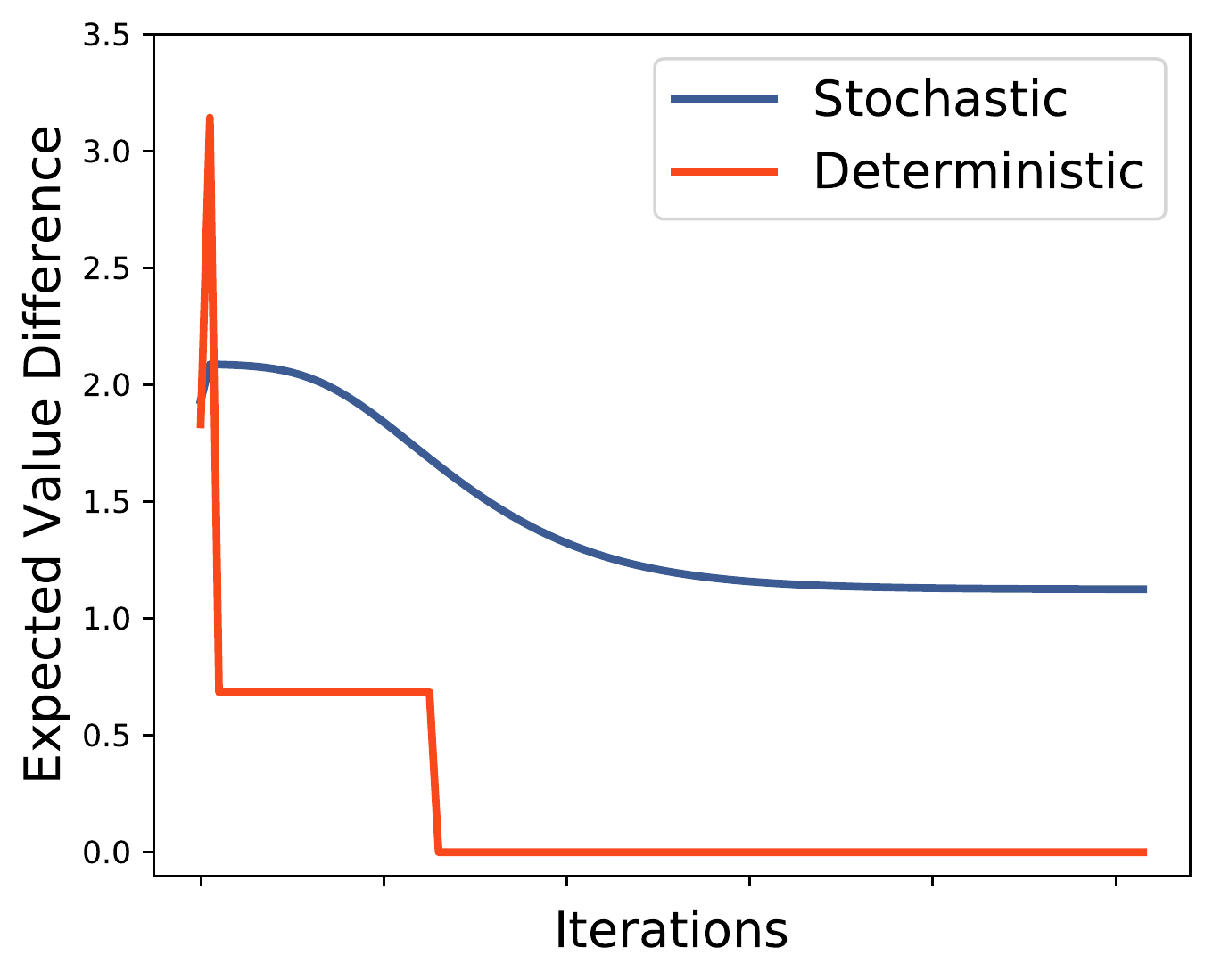}} &
    \MSHangBox{\includegraphics[width=0.25\textwidth]{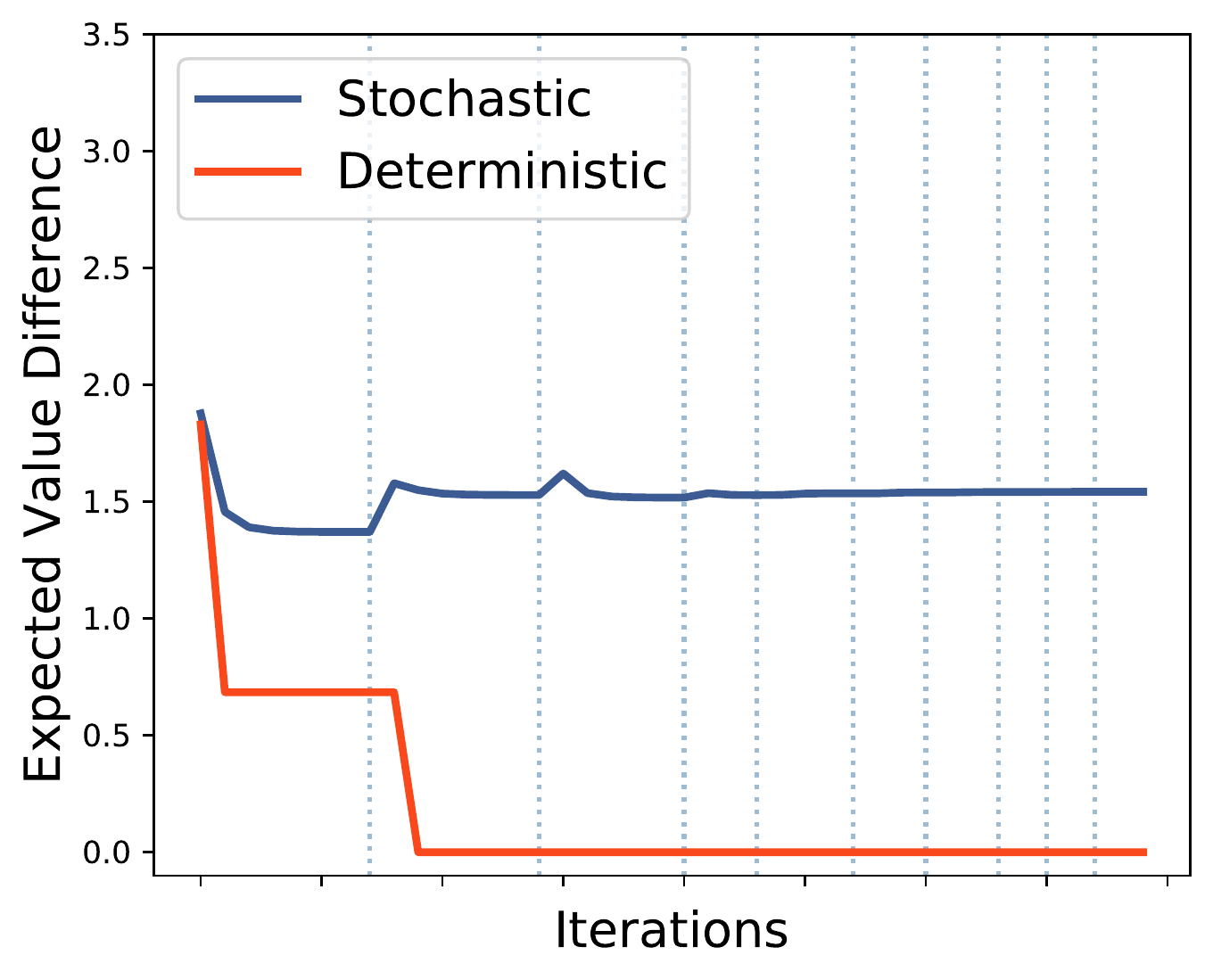}} \\
    ``Corridor" Expert Policy & MCE-IRL & NPL ($K=10$) \\
    
    \MSHangBox{\includegraphics[width=0.2\textwidth]{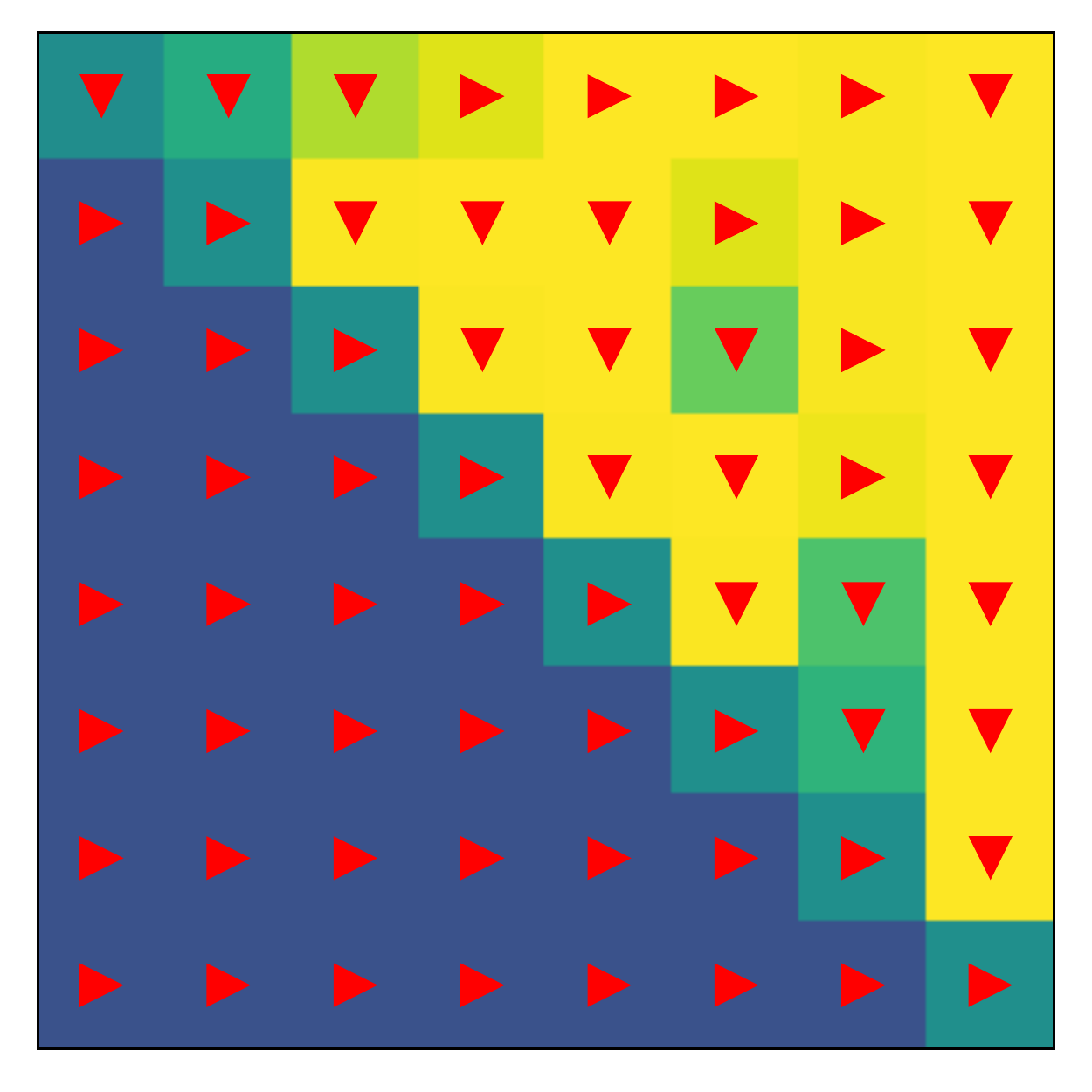}}&
    \MSHangBox{\includegraphics[width=0.25\textwidth]{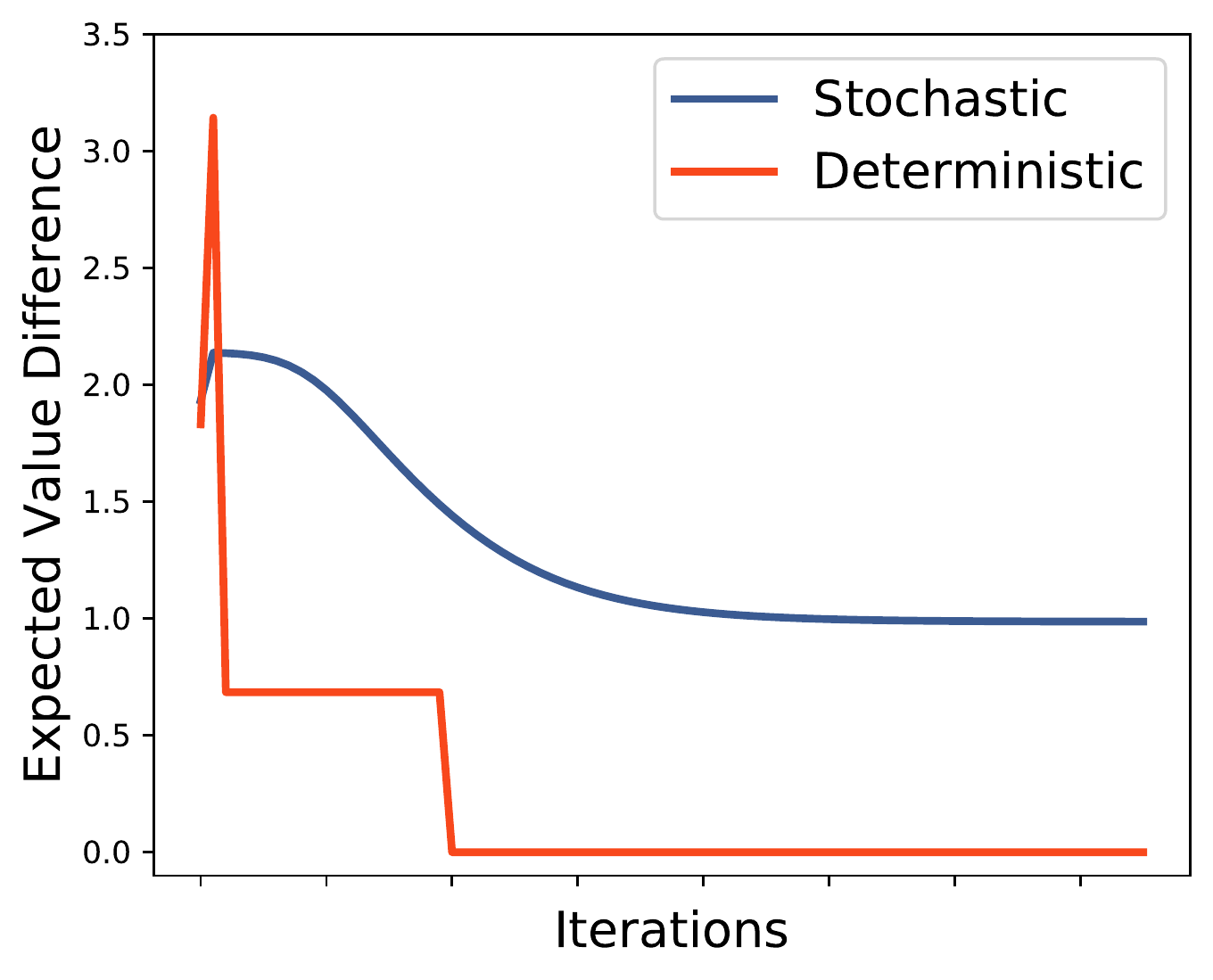}} &
    \MSHangBox{\includegraphics[width=0.25\textwidth]{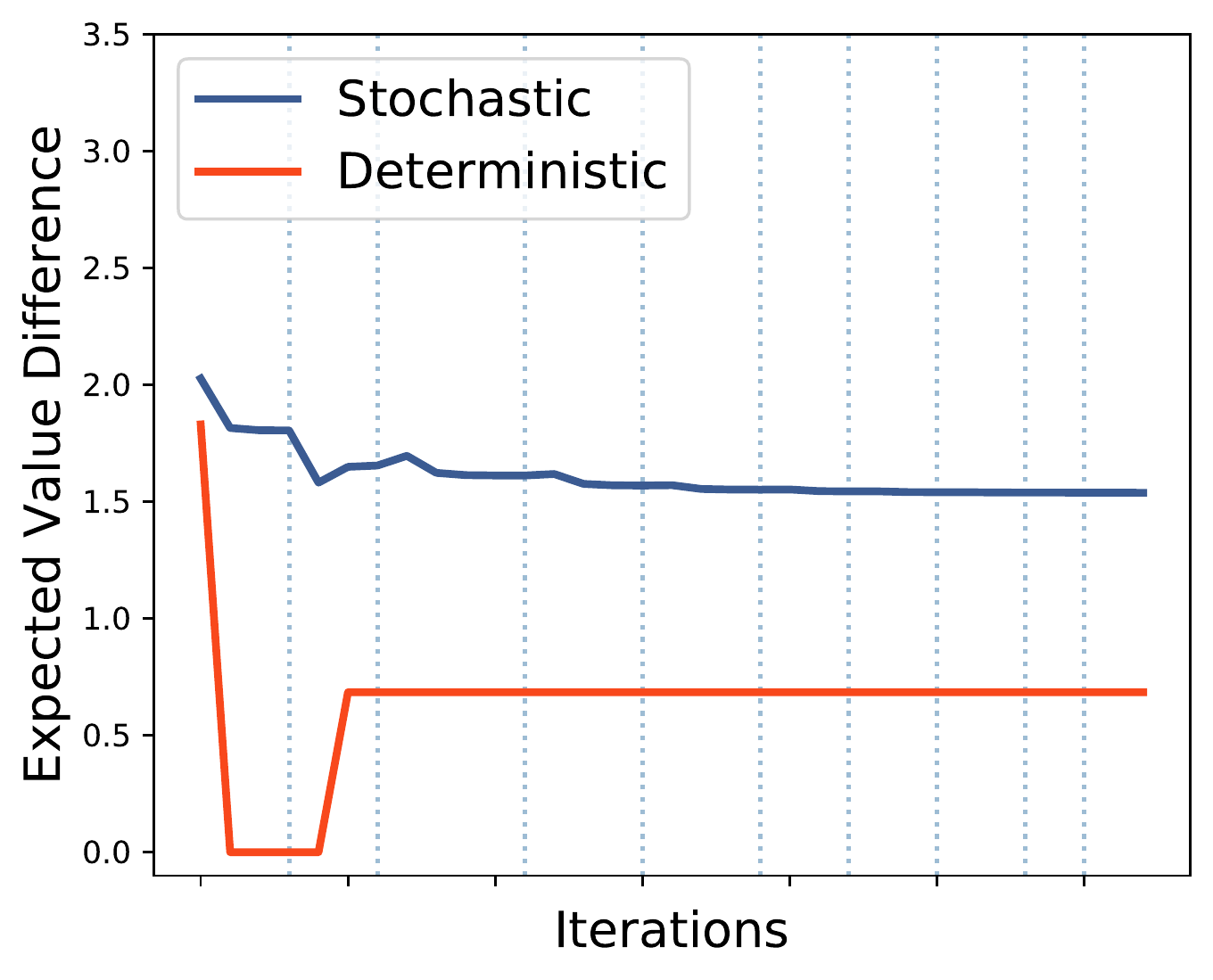}} \\
    ``Triangle" Expert Policy & MCE-IRL & NPL ($K=10$) \\ 
    \end{tabular}
    
    \caption{Partial expert policy estimates and the algorithms' performances (EVD). Warmer colors indicate more deterministic policies in the direction of the red arrows. Blue cells indicate uniform random policy. The right two columns show algorithms' stochastic and deterministic EVD performance. Vertical lines in the NPL plots indicate where $\tilde{\pi}$ was updated ($K=10$). }
    \label{fig:img_policy_evd_obstaclegrid}
\end{figure*}

In Section~\ref{subsec:feat}, we observed that NPL does not perform well in low-data regimes. This is reasonable since the NPL (and CCP) method depends on expert policy estimates to decide a state's relative importance, as well as to approximate the optimal soft value (\ref{eqn:nfxp-policy}, \ref{eqn:ccp-gradient}). While we previously experimented with ``low data'' that was not specific to parts of the state space, in this section, we experiment with a more specific lack of data, in order to investigate the effect of the estimate of a state's relative importance on the performance of NPL.

\noindent {\bf Obstacleworld performance:} To see the effect of relative state importance we use the Obstacleworld environment used in Section~\ref{subsec:feat} and compare results when we only have partial expert policy estimates, \emph{i.e.,} only in certain parts of the state space. Figure~\ref{fig:img_policy_evd_obstaclegrid} (left row) plots the two configurations we use for this experiment. The yellow region in each plot shows states that we have the expert policy estimates for while the blue region indicates states for which we do not have any expert policy estimates. 
Thus, for the top row we do not have expert policy estimates in the middle corridor (diagonal) while for the bottom row we do not have expert policy estimates for any state below the  corridor (diagonal).

Figure~\ref{fig:img_policy_evd_obstaclegrid} (right) compares the EVD and stochastic-EVD for both of these settings between MCE-IRL and NPL. In both cases, MCE-IRL performs well, as the feature counts (\emph{i.e.,} path counts) are well-estimated. 
We observe that when the environment only lacks the expert policy estimates along the corridor (top row), the NPL method is able to recover the true reward and performs as well as MCE-IRL. On the other hand, when we lack the expert policy estimates for the bottom half of the grid, the NPL method performs poorly and is not able to recover the reward at all. 
This is because states along the corridor (diagonal) are less important since we already have expert policy estimates for states slightly away from the diagonal (Figure~\ref{fig:img_policy_evd_obstaclegrid} top row). Thus even if the expert takes a sub-optimal action in the corridor states, the expert policy estimates near the corridor will still lead the agent to high-reward state space regions, \emph{i.e.} towards the goal.
In contrast, if we do not have any expert policy estimates for the bottom half of the grid (Figure~\ref{fig:img_policy_evd_obstaclegrid} bottom row) the agent cannot figure out the high reward region from any of the states in this region. Given this complete lack of expert policy estimates for an important set of states the NPL method performs poorly as compared to MCE-IRL.

Through our experiments with Obstacleworld, we have demonstrated that, since the NPL gradient \eqref{eqn:ccp-gradient} depends on the estimate of a state's relative importance, the performance of NPL is highly dependent on which part of the state space lacks data, as well as the amount of data available.

\section{Experimental Details}
\label{app:D}
In this section, we discuss our experimental setup, qualitative results, and insights about the EPIC metric.

\subsection{Experimental Setup}
\noindent {\bf Transition function: } In Obstacleworld, the transition dynamics was deterministic. Objectworld has stochastic dynamics with a probability 0.3 of moving the agent to a random neighboring grid. For MountainCar, we estimated the transition function for continuous $(s',s)$ from expert trajectories using kernels: $T(s'|s,a) = \frac{\sum_i \;\; \mathbf{I}(a_i = a) \phi_i(s) \psi_i(s')}{\sum_i \;\; \mathbf{I}(a_i = a) \phi_i(s)}$. Here, $i$ indexes expert data, $a_i$ is the action taken in the $i^\text{th}$ data point, and $\phi_i, \psi_i$ are Gaussian kernels. The state space was then discretized by $100$ for position and velocity.

\noindent {\bf Convergence criteria: } For fair comparison of both performance and time measurement across different methods, we define a \emph{common} convergence criterion. Specifically, we consider convergence when the gradient norm falls below a certain threshold. Since deep neural networks can often have noisy gradients, for non-linear reward settings, we consider convergence when the gradient norm is below a certain threshold for multiple iterations. 
For NPL, this is the convergence criterion for the inner loop ($\bt$ update) at all $K$. The outer loop ($\tilde{\pi}$ update) is run for a fixed $K$. In Obstacleworld and MountainCar, we set $K=10$ while in Objectworld, we set $K=5$. Empirically, we observe that $\tilde{\pi}$ does not change after $K=3$.

\noindent {\bf Expert trajectories: } For each environment, we first compute a policy under the true reward then use the policy to generate expert trajectories. In Obstacleworld and MountainCar, we first compute the optimal policy in the soft value sense and sample expert trajectories using this policy. Adding stochasticity to the optimal policy allows us to empirically investigate convergence properties under varying expert conditions. For ObjectWorld, we compute the optimal policy and sample expert trajectories using this policy. Then, we add stochastic noise with probability 0.3 to the optimal policy.
For Obstacleworld, the number of expert trajectory was $\in \{1, 3, 5, 10, 20, 30, 50\}$. Since our experiments with MountainCar considered a large state space, the number of expert trajectories was $\in \{100, 200, 400, 500, 700, 1000\}$. For Objectworld, the number of expert trajectories was $\in \{1, 3, 5, 10, 20, 30, 50, 100, 150, 200\}$. 

\noindent {\bf Seeds:} Results for each experiment were generated by running each algorithm with multiple seeds. For use 3 seeds for Obstacleworld and MountainCar while we use 5 seeds for Objectworld.

\noindent {\bf Implementation: } To fairly compare the training time between methods we use vectorized implementations throughout. 
For deep reward representations we use PyTorch \cite{NEURIPS2019_9015}. Our reward network is a two-layer feed-forward neural network with size $[32, 16]$ respectively, and ReLU non-linearities. The network was trained with Adam optimizer.

\noindent {\bf Platform: } All experiments were executed on Linux Ubuntu 18.04 platform equipped with Intel Xeon CPU E5-2690 v4. We did not use GPUs because we observed that our implementation was faster on CPU only. When measuring the training time, we were careful not to run redundant processes to have a fair and precise measurement.

\begin{figure}[t!]
    \centering
    \includegraphics[width=0.8\columnwidth]{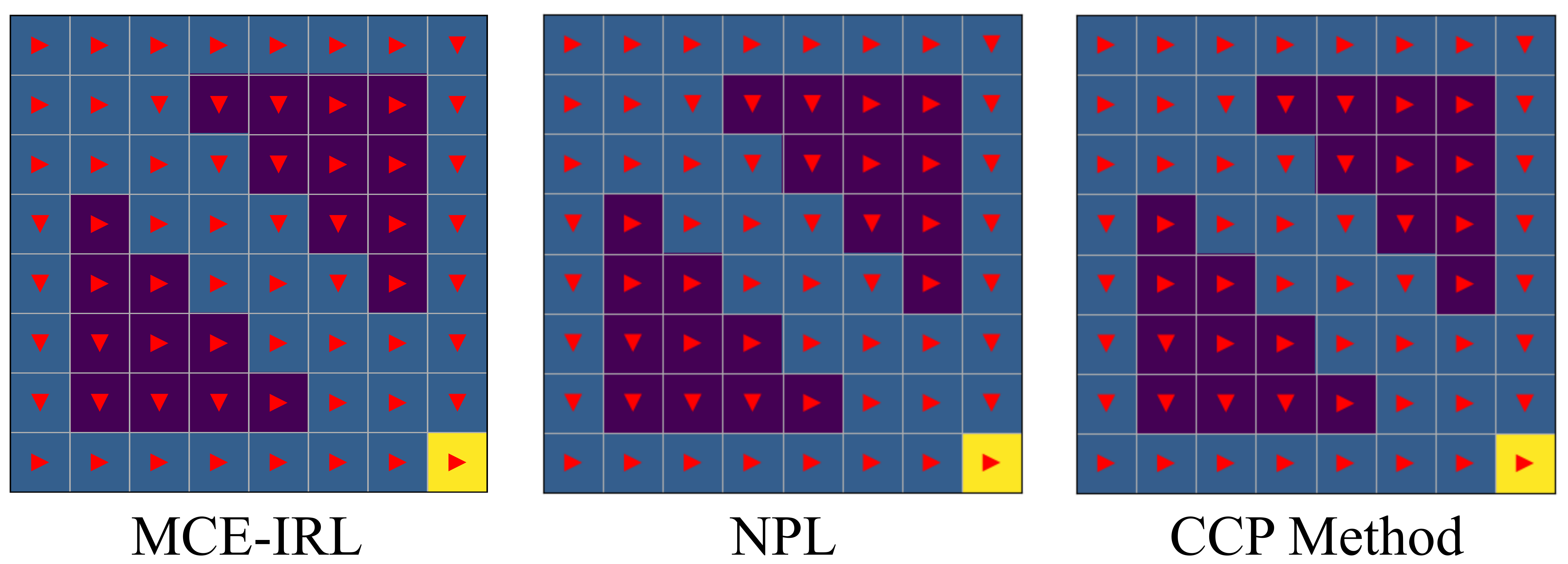}
    \caption{  \footnotesize{{\bf Obstacleworld:} Recovered reward with 50 expert trajectories as input. 
    }}
    \label{fig:img_performance_obstacleworld_qualitative}
\end{figure}

\begin{figure}[t!]
    \centering
    \includegraphics[width=0.8\columnwidth]{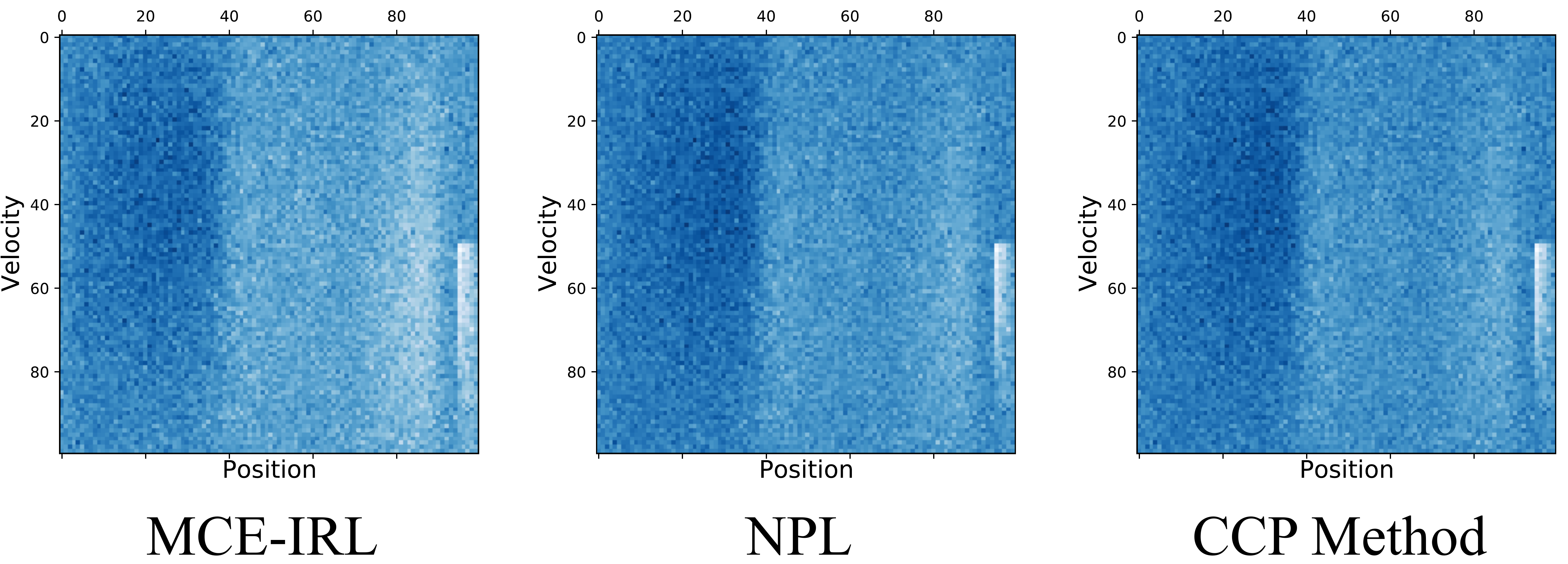}
    \caption{  \footnotesize{{\bf MountainCar:} Recovered reward with 1000 expert trajectories as input. 
    }}
    \label{fig:img_performance_mountaincar_qualitative}
\end{figure}

%

\subsection{Qualitative Results}
In our paper, we provided qualitative results for the Objectworld environment (Figure~\ref{fig:img_performance_objectworld_qualitative}). We further show the qualitative result for the Obstacleworld and MountainCar environment in Figure~\ref{fig:img_performance_obstacleworld_qualitative} and \ref{fig:img_performance_mountaincar_qualitative} respectively. As discussed in Section~\ref{sec:experiments}, when any method takes large amount of data as input, the qualitative results showed that they converge to similar metric values. Qualitative results show that all methods result in similar reward, thus supporting quantitative results.

\subsection{Insight of the EPIC metric}

For all environments we see that the EPIC metric shows improving performance with more data. Since higher data regimes allow each method to better approximate the true reward, this indicates that EPIC is able to express a distance metric between reward functions. However, in Figure~\ref{fig:img_performance_objectworld} (top-right) we see that EPIC for the CCP-method is \emph{much lower} than for other methods. 
While this might seem like an anomaly, we argue that this, in fact, is a limitation of the EPIC metric.
Specifically, since EPIC directly compares inferred rewards without computing a policy,
it may incorrectly distinguish between different rewards (\emph{i.e.} rewards that do not belong to the equivalence class used in \cite{gleave2020quantifying}) but which result in similar optimal policies $\pi_{\bt^*}$.
We see empirical evidence of this in  Figure~\ref{fig:img_performance_objectworld} (middle-figures), which shows that both EVD and stochastic-EVD are similar for all methods. This indicates that the policies recovered by all methods using the respectively recovered rewards are quite similar.

\end{appendices}
\end{document}